\documentclass[conference]{IEEEtran}
\IEEEoverridecommandlockouts
\usepackage{cite}
\usepackage{amsmath,amssymb,amsfonts,amsthm}
\usepackage{algorithmic}
\usepackage{graphicx}
\usepackage{textcomp}
\usepackage{xcolor}
\usepackage{booktabs}
\usepackage{multirow}
\usepackage{subfigure}
\usepackage{wrapfig}
\usepackage{hyperref}
\usepackage{amsmath, amssymb}
\makeatletter
 \newcommand{\linebreakand}{%
      \end{@IEEEauthorhalign}
      \hfill\mbox{}\par
      \mbox{}\hfill\begin{@IEEEauthorhalign}
    }
\makeatother
\newtheorem{theorem}{Theorem}

\newcommand{\bst}[1]{{\textbf{\textcolor{red}{#1}}}}
\newcommand{\subbst}[1]{\textcolor{blue}{\underline{{#1}}}}
\newcommand{\scalea}[1]{\scalebox{0.78}{#1}}
\newcommand{\scaleb}[1]{\scalebox{0.8}{#1}}
\usepackage{amsmath}
\usepackage{xspace}

\usepackage{caption}

\usepackage{listings}
\def\BibTeX{{\rm B\kern-.05em{\sc i\kern-.025em b}\kern-.08em
    T\kern-.1667em\lower.7ex\hbox{E}\kern-.125emX}}
\definecolor{abl}{HTML}{74709b}

\begin{document}
\title{
FSMLP: Modelling Channel Dependencies With Simplex Theory Based Multi-Layer Perceptions In Frequency Domain

}

\author{
\IEEEauthorblockN{Zhengnan Li}
\IEEEauthorblockA{\textit{Communication University of China} \\
Beijing, China \\
lzhengnan389@gmail.com}
\and
\IEEEauthorblockN{Haoxuan Li}
\IEEEauthorblockA{\textit{Peking University} \\
Beijing, China}
\and
\IEEEauthorblockN{Hao Wang}
\IEEEauthorblockA{\textit{Zhejiang University} \\
Hangzhou, China } 
\linebreakand
\IEEEauthorblockN{Jun Fang}
\IEEEauthorblockA{\textit{Tsinghua University}\\
Beijing, China }
\and
\IEEEauthorblockN{Zhichao Chen}
\IEEEauthorblockA{\textit{Peking University} \\
Beijing, China }
\and
\IEEEauthorblockN{Yuting Tan}
\IEEEauthorblockA{\textit{Communication University of China} \\
Beijing, China }
\and
\IEEEauthorblockN{Xilong Cheng}
\IEEEauthorblockA{\textit{Communication University of China} \\
Beijing, China }
\and
\IEEEauthorblockN{Yunxiao Qin}
\IEEEauthorblockA{\textit{Communication University of China} \\
Beijing, China \\
qinyunxiao@cuc.edu.cn}
}



\maketitle

\begin{abstract}
Time series forecasting (TSF) plays a crucial role in various domains, including web data analysis, energy consumption prediction, and weather forecasting. While Multi-Layer Perceptrons (MLPs) are lightweight and effective for capturing temporal dependencies, they are prone to overfitting when used to model inter-channel dependencies. 
In this paper, we investigate the overfitting problem in channel-wise MLPs using Rademacher complexity theory, revealing that extreme values in time series data exacerbate this issue.
To mitigate this issue, we introduce a novel Simplex-MLP layer, where the weights are constrained within a standard simplex. This strategy encourages the model to learn simpler patterns and thereby reducing overfitting to extreme values. 
Based on the Simplex-MLP layer, we propose a novel \textbf{F}requency \textbf{S}implex \textbf{MLP} (FSMLP) framework for time series forecasting, comprising of two kinds of modules: \textbf{S}implex \textbf{C}hannel-\textbf{W}ise MLP (SCWM) and \textbf{F}requency \textbf{T}emporal \textbf{M}LP (FTM). 
The SCWM effectively leverages the Simplex-MLP to capture inter-channel dependencies, while the FTM is a simple yet efficient temporal MLP designed to extract temporal information from the data.
Our theoretical analysis shows that the upper bound of the Rademacher Complexity for Simplex-MLP is lower than that for standard MLPs.
Moreover, we validate our proposed method on seven benchmark datasets, demonstrating significant improvements in forecasting accuracy and efficiency, while also showcasing superior scalability.
Additionally, we demonstrate that Simplex-MLP can improve other methods that use channel-wise MLP to achieve less overfitting and improved performance.
Code are available \href{https://github.com/FMLYD/FSMLP}{\textcolor{red}{here}}.


\end{abstract}


\section{Introduction}

Time series forecasting (TSF) is crucial across various fields, including web data analysis \cite{wu2021autoformer}, electricity consumption \cite{ECL, foundationsurvey, lstn}, and weather forecasting \cite{itransformer, liu2024timer, convtimenet, tft}. Accurate predictions from historical data are essential for decision-making, policy development, and strategic planning. 
Recent advances in deep learning have significantly enhanced TSF capabilities \cite{xue2024card, liu2023nonstationary, oreshkin2020nbeats, moderntcn,onefitsall,yi2024filternetharnessingfrequencyfilters,preformer,difformer}.

\begin{figure}
    \centering
    \includegraphics[width=\linewidth]{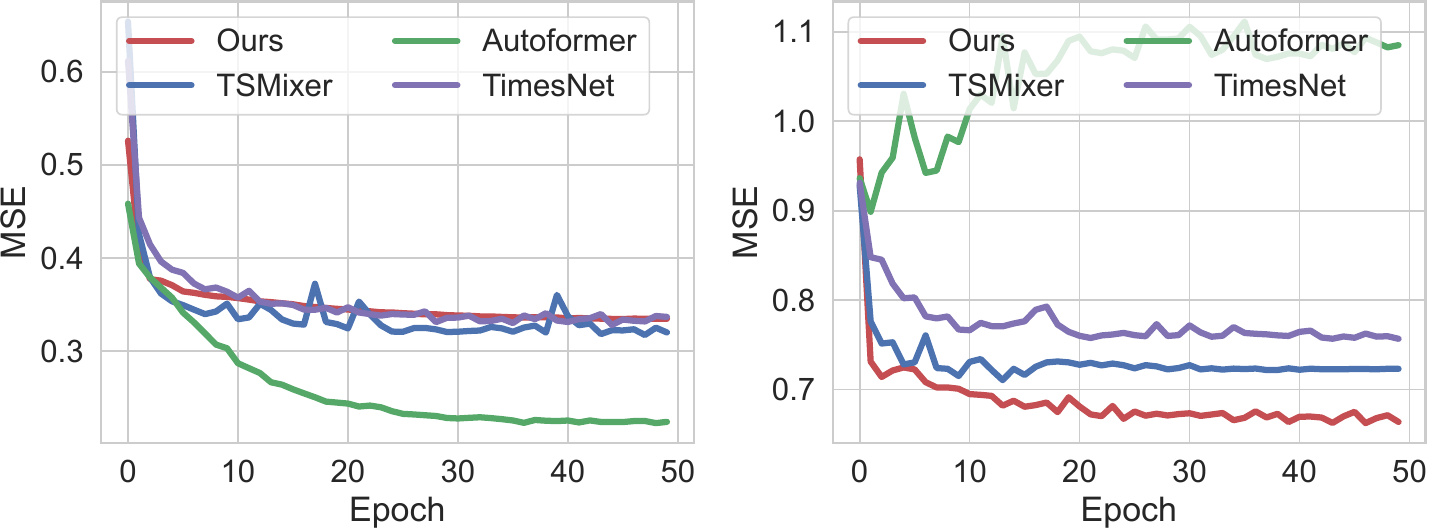}
    \caption{Comparison of overfitting behavior among FSMLP, TimesNet, TSMixer, and Autoformer on the ETTh1 dataset, with both look-back window and prediction length set to 96. TSMixer, TimesNet, and Autoformer show a rapid decline in training loss, but their validation loss remains high, indicating overfitting. In contrast, FSMLP maintains lower validation loss, demonstrating that the Simplex MLP effectively mitigates the overfitting issues typically encountered by MLPs when modeling channel dependencies.}
    \label{fig:loss}
\end{figure}

Deep learning-based TSF methods can be categorized into channel-independent and channel-mix approaches. Channel-independent methods \cite{dlinear, fits} focus on capturing temporal dependencies by using a shared model across different channels. These methods often exhibit robust performance. For example, PatchTST \cite{patchtst} segments time series into patches and employs attention mechanisms to capture relationships between these patches, thereby effectively modeling temporal dependencies. 
However, the absence of explicit inter-channel modeling can result in suboptimal performance when the relationships between channels contain valuable information \cite{itransformer, moderntcn,samformer}. 

Unlike channel-independent methods, channel-mix models excel at capturing inter-channel relationships by explicitly modeling the dependencies among channels. 
There exist various approaches to model these dependencies, among which one of the simplest and most intuitive methods is the use of MLPs (Multi-Layer Perceptrons).
This approach has been employed in early Transformer models \cite{fedformer,wu2021autoformer,liu2023nonstationary,informer,ni2024basisformer}, such as Fedformer \cite{fedformer}, Autoformer \cite{wu2021autoformer}, Nonstationary-Transformer \cite{liu2023nonstationary}, and MLP-Mixer-like models such as TSMixer \cite{chen2023tsmixerallmlparchitecturetime}. 
However, despite success in modelling temporal dependencies,these channel-wise MLP models  often suffer from overfitting and display sub-optimal performance compared to channel-wise attention methods  \cite{itransformer}.
As shown in Fig. \ref{fig:loss}, TimesNet, TSMixer, and Autoformer all exhibit signs of overfitting.

In this paper, \textbf{we leverage Rademacher complexity theory~\cite{rademacher} to analyze this phenomenon and find that this overfitting issue may be attributed to the presence of extreme values in time series data}, as shown in Table \ref{tab:zero}. 
Rademacher complexity measures the ability of a function class to fit random noise, with lower Rademacher complexity indicating a reduced tendency to overfit. 
The Rademacher complexity \(\mathcal{R}_S(\mathcal{H})\) for the hypothesis class \(\mathcal{H}\) of the MLP in linear regression is bounded as follows:
\(
\mathcal{R}_S(\mathcal{H}) \leq \frac{B}{m} \sqrt{\sum_{i=1}^m \| x^{(i)} \|_2^2}
\)
where \( w \in \mathbb{R}^d \) are the weight parameters of the model, \( d \) is the input dimensionality, \( m \) is the number of training data points, \( x^{(i)} \in \mathbb{R}^d \) represents the \( i \)-th input data point, \( \| x^{(i)} \|_2 \) is the \( \ell_2 \)-norm of the \( i \)-th data point, and \( B \) is an upper bound on the norm of the weight vector \( w \), typically assumed to be \( B = \| w \|_2 \) and much greater than 1.
The presence of extreme values in time series data can result in a large \( B \) when modeling channel dependencies with MLPs, thereby increasing the Rademacher complexity and making these models more prone to overfitting.


\begin{table}
    \centering
    \caption{In mainstream time series datasets, extreme values are observed, with the majority of values falling within \( \sigma \), while a few outliers exceed \( 3\sigma \). These extreme values significantly impact the performance of MLPs in capturing channel dependencies.}
        \label{tab:zero}
    \renewcommand{\arraystretch}{1.2}
    \resizebox{\linewidth}{!}{\begin{tabular}{c|ccccccc}
    \toprule
         &  ETTh1&  ETTh2&  ETTm1&  ETTm2&  Traffic&  Weather& ECL\\
         \midrule
        $\leq \sigma$ &  87.47\%&  87.22&  12.46\%&  87.30\%&  88.44\%&  86.69\%& 84.38\%\\
        $\geq 3\sigma$ &  0.35\%&  0.99\%&  0.35\%&  0.99\%&  1.59\%&  0.8\%& 0.39\%\\
        \bottomrule
    \end{tabular}}
    
\end{table}

To address the overfitting oriented from the extreme values, \textbf{we propose a novel operator: Simplex-MLP, inspired by the theory of the Standard \(n\)-simplex} \cite{simplex}.
A standard \(n\)-simplex is defined as the set of points in \(\mathbb{R}^{n+1}\) that satisfy two conditions: the sum of the coordinates of each point equals $1$, and each coordinate is greater than or equal to zero. 
We apply the \(n\)-simplex theory to traditional MLP and constrain the weights of MLP to lie within the standard \(n\)-simplex.
We call such a novel MLP as Simplex-MLP and the Rademacher complexity of it can be bounded by 
\(
\mathcal{R}_S(\mathcal{H}_\Delta) \leq \frac{1}{m} \sqrt{\sum_{i=1}^m \| x^{(i)} \|_2^2}
\), much smaller than that of traditional MLPs.
This indicates that the Simplex-MLP reduces the influence of redundant noise among channels and thereby improving generalization and reducing overfitting.

Based on Simplex-MLP, we introduce a novel channel-mix framework \textbf{F}requency \textbf{S}implex \textbf{MLP} (FSMLP) for time series forecasting.
Specifically, FSMLP is composed of two kinds of modules: Simplex Channel-Wise MLP (SCWM) and Frequency Temporal MLP (FTM). SCWM can effectively extract information among channels, while FTM excels at extracting temporal information within each channel.

Furthermore, our models capture both temporal and inter-channel dependencies in the frequency domain, enhancing the overall performance.
Each frequency component represents a period in the time domain. Therefore, modeling inter-channel dependencies in the frequency domain involves modeling the dependencies between different periods  across channels. This approach introduces less noise compared to directly modeling inter-channel dependencies in the time domain\cite{atfnet,yi2024filternetharnessingfrequencyfilters,frets,ftmixer}.

In summary, our contributions are as follows:
\begin{itemize}
    \item We utilized Rademacher complexity to analyze and identify that the use of MLPs for explicitly modeling inter-channel dependencies leads to overfitting, primarily due to the presence of extreme values in time series dataset.
    
    \item To mitigate overfitting, we introduce a novel Simplex-MLP layer, which constrains the weights of the MLP to lie within a well-defined standard \(n\)-simplex, making the MLP less prone to overfitting. Besides, based on Simplex-MLP, we proposed a novel framework FSMLP for time series forecasting.
    
    \item Experimentally, we \textbf{achieve much better performances than existing state-of-the-art methods on seven popularly used benchmarks, with significant margins}, demonstrates the effectiveness of the proposed Simplex-MLP layer and FSMLP framework.
    Moreover, through experiments, we also demonstrate that the proposed Simplex-MLP can also greatly improve other methods.
\end{itemize}

\vspace{10pt}
\section{Related Work}

\subsection{Time Series Forecasting}

Time series forecasting (TSF) is an fundamental task with wide-ranging applications in various domains, such as web data analysis \cite{wu2021autoformer}, energy consumption \cite{ECL, foundationsurvey, lstn,film,fre}, and weather prediction \cite{itransformer, liu2024timer, convtimenet, tft}.  
The development of deep learning has significantly transformed the landscape, introducing models that can capture complex temporal dependencies and nonlinear patterns in the data. 
Recent innovations, particularly in Transformer-based models, have pushed the boundaries of TSF performance \cite{xue2024card, liu2023nonstationary, oreshkin2020nbeats, moderntcn}. Models like Informer \cite{informer} and Autoformer \cite{wu2021autoformer} have demonstrated remarkable success by effectively handling long-term dependencies and scalability issues.
On the other hand MLP-based methods also has its place,  known for their light-weight and effectiveness, some of them can even surpass some complex Transformer-based models with one or two single linear layers. DLinear\cite{dlinear} decompose the time series into season and trend,  RLinear\cite{revin} propose a instance normalization then use a linear model to learn temporal patterns. 

\subsection{Channel-Independent and Channel-Mix Methods}

Deep learning-based TSF methods can be broadly categorized into channel-independent and channel-mix methods. 
Channel-independent methods\cite{tide,dlinear,revin,patchtst,fredo,atfnet}, such as DLinear \cite{dlinear} and RLinear \cite{revin}, focus on modeling each time series channel with a shared model. These methods generally exhibit robust performance as the input sequence length increases, but not modeling inter-channel dependencies.  for example PatchTST \cite{patchtst}, segments time series into patches and employs attention mechanisms to capture relationships within these patches, achieving significant success without considering inter-channel correlations.

Channel-mix models\cite{fredf,liu2024mamba4rec,mamba,yang2024uncovering,itransformer,moderntcn}, on the other hand, aim to capture dependencies between different channels.
It's intuitive to capture channel-dependencies by MLPs.
This approach has been employed in early Transformer models \cite{fedformer,wu2021autoformer,liu2023nonstationary,informer,oreshkin2020nbeats}, such as Fedformer \cite{fedformer}, Autoformer \cite{wu2021autoformer}, Nonstationary-Transformer \cite{liu2023nonstationary}, and MLP-Mixer-like models such as TSMixer \cite{chen2023tsmixerallmlparchitecturetime}. TSMixer \cite{chen2023tsmixerallmlparchitecturetime}, for instance, models inter-channel dependencies using MLPs in the time domain, while FreTS \cite{frets} utilizes complex-valued MLPs to capture these dependencies in the frequency domain. Despite these innovations, MLP-based models still face challenges related to overfitting, similar to early  methods.
Thus, there is a need to develop an enhanced MLP that efficiently captures channel dependencies while maintaining high performance.

\subsection{Frequency Domain Approaches}

The frequency domain offers a powerful perspective for analyzing time series data, enabling the identification and capture of global dependencies that might be less apparent in the time domain \cite{fits, atfnet, yi2024filternetharnessingfrequencyfilters}. Transforming time series data into the frequency domain using techniques like the Discrete Fourier Transform (DFT) allows models to focus on periodic patterns and long-term dependencies. FreTS \cite{frets} exemplifies this approach by leveraging complex-valued MLPs in the frequency domain to model global dependencies, resulting in improved performance. The global nature of frequency domain representations is particularly beneficial for capturing inter-channel dependencies, as each frequency component corresponds to a periodicity, thus enabling a more holistic modeling of relationships across different channels.
However, FreTS does not explicitly model inter-channel dependencies using MLPs. Specifically, FreTS first applies word embedding techniques to embed different time stamps across channels, and then performs frequency domain transformations along the time and channel dimensions. Afterward, MLPs are applied to the embeddings. This lack of explicit  modeling for inter-channel dependencies and temporal dependencies limits the model's performance.

\vspace{10pt}
\section{Preliminaries}





\subsection{Standard \( n \)-Simplex}
Constraining parameters to lie within the standard \( n \)-simplex is a technique in machine learning to ensure non-negativity and boundedness \cite{simplex}, which can help prevent overfitting.
To constrain the weights \( \mathbf{W} \in \mathbb{R}^{n \times d} \) of a model to lie in the standard \( n \)-simplex.

An \( n \)-simplex \cite{simplex} is a generalization of the notion of a triangle or tetrahedron to \( n \) dimensions. It is defined as the convex hull of its \( n+1 \) vertices in \( \mathbb{R}^n \). Specifically, the \( n \)-simplex is the set of points \( \mathbf{w} \) such that:
\begin{equation}
\Delta^n = \left\{ \mathbf{w} \in \mathbb{R}^n \mid \mathbf{w} = \sum_{i=0}^{n} \lambda_i \mathbf{v}_i, \sum_{i=0}^{n} \lambda_i = 1, \lambda_i \geq 0 \right\},
\end{equation}
where \( \mathbf{v}_i \in \mathbb{R}^{n+1} \)is the $i$-th vertice of the simplex.

In standard \( n \)-simplex, each $i$-th vertice \( \mathbf{v}_i \)  is a standard basis vector. Formally, the standard \( n \)-simplex \( \Delta^n \) is given by:

\begin{equation}
\Delta^n = \left\{ \mathbf{w} \in \mathbb{R}^{n+1} \mid \sum_{i=0}^{n} w_i = 1 \text{ and } w_i \geq 0 \text{ for all } i \right\}
\end{equation}

This can be visualized as the convex hull of the \( n+1 \) standard basis vectors \( \mathbf{e}_i \) in \( \mathbb{R}^{n+1} \), where:

\begin{equation}
\mathbf{e}_i = (0, \ldots, 0, 1, 0, \ldots, 0) \quad \text{for } i = 0, 1, \ldots, n
\end{equation}

\subsection{Problem Definition}
Let $[ {X}_1, {X}_2, \cdots, {X}_T ] \in \mathbb{R}^{N \times T}$ stand for the regularly sampled multi-channel time series dataset with $N$ series and $T$ timestamps, where ${X}_t \in \mathbb{R}^N$ denotes the multi-channel values of $N$ distinct series at timestamp $t$.
We consider a time series lookback window of length-$L$ at each timestamp $t$ as the model input, namely $\mathbf{X}_t = [{X}_{t-L+1}, {X}_{t-L+2}, \cdots, {X}_{t} ] \in \mathbb{R}^{N \times L}$; also, we consider a horizon window of length-$\tau$ at timestamp $t$ as the prediction target, denoted as $\mathbf{Y}_t = [{X}_{t+1}, {X}_{t+2}, \cdots, {X}_{t+\tau} ] \in \mathbb{R}^{N \times \tau}$. 
Then the time series forecasting formulation is to use historical observations $\mathbf{X}_t$ to predict future values ${\mathbf{Y}}_t$.
For simplicity, we shorten the model input $\mathbf{X}_t$ as $\mathbf{X} = [{X}_{1}, {X}_{2}, \cdots, {X}_{L} ] \in \mathbb{R}^{N \times L}$ and reformulate the prediction target as $\mathbf{Y} = [{X}_{L+1}, {X}_{L+2}, \cdots, {X}_{L+\tau} ] \in \mathbb{R}^{N \times \tau}$, in the rest of the paper.

\vspace{10pt}
\section{Methodology}
In this section, we first provide the details of Simplex-MLP and then describe the architecture of the proposed FSMLP.
\begin{figure*}
    \centering
    \includegraphics[width=\linewidth]{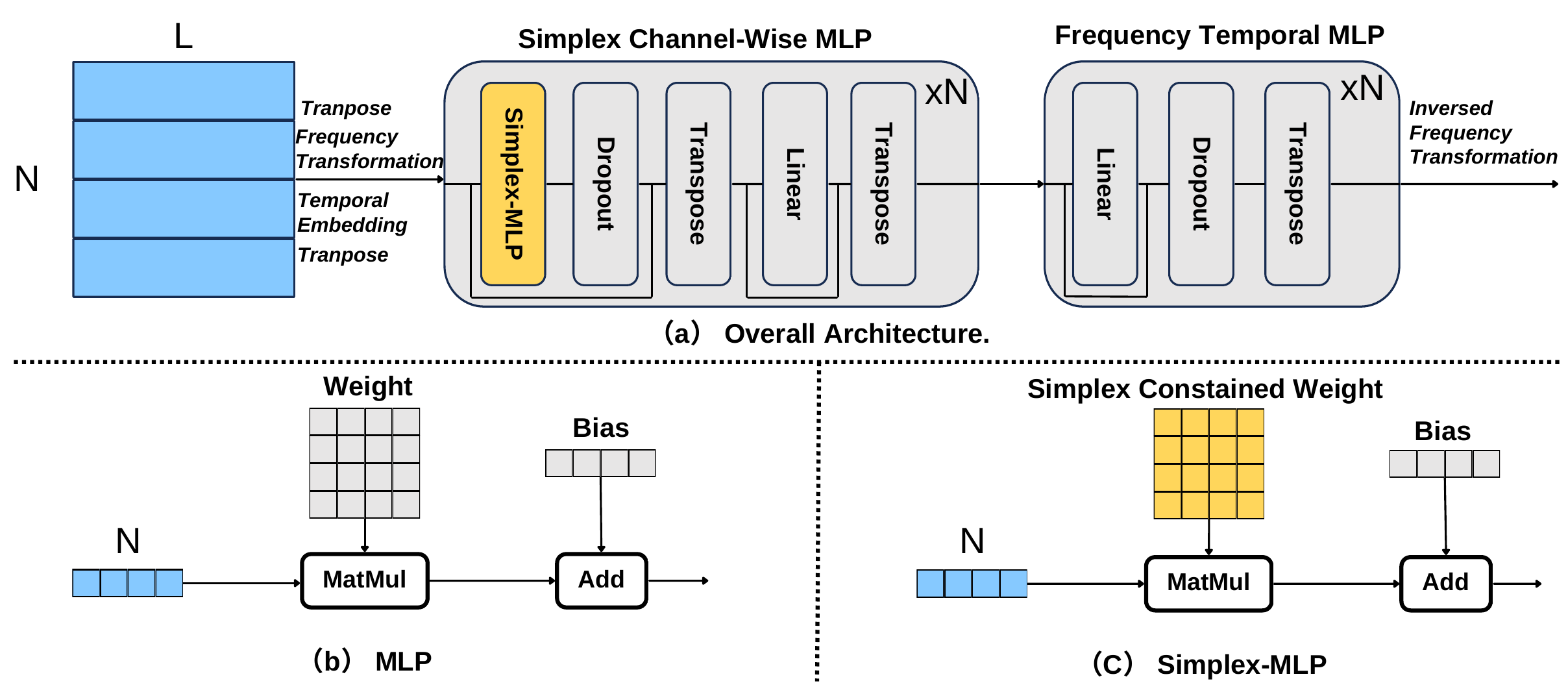}
    \caption{The overall architecture of the proposed FSMLP. We first extract the inter-channel dependencies with our Simplex Channel-Wise MLP then use the Frequency Temporal MLP to extract temporal dependencies.}
    \label{fig:model}
\end{figure*}

\subsection{Simplex-MLP}
\label{sec:simplex-mlp}
Traditional MLP is commonly formulated as:
\begin{equation}
X_{\text{out}} = \text{Matmul}(X_{\text{in}}, W) + b,
\end{equation}
where the weights $W$ are unconstrained.
This can lead to large or unbounded values that overfit the data, particularly in high-dimensional settings. 
To address the overfitting problem commonly encountered in traditional linear layers when modeling complex channel dependencies, we introduce the \textbf{Simplex-MLP} layer.
The key motivation behind this approach is the imposition of a geometric constraint on the weight space, specifically ensuring that the weights reside within the \textbf{Standard N-Simplex}. This constraint naturally bounds the weights within a well-defined region, promoting a more compact and structured representation of the data, which helps reduce Rademacher complexity.
We formulate the Simplex MLP as:
\begin{equation}
X_{\text{out}} = \text{Matmul}(X_{\text{in}}, f_{\text{sim}}(W)) + b,
\end{equation}
where $f_{\text{sim}}$ is the function to constrain weights $W$ to lie on the {Standard N-Simplex}.
Both the proof in Section~\ref{sec:proof} and the experimental results in Section \ref{sec:Analysis of Simplex-MLP Improvements} verify that based on the proposed Simplex MLP, the model becomes less susceptible to memorizing spurious correlations or noise in the data, leading to significantly reduced overfitting.



The $f_{\text{sim}}$ function can be detailed as $f_{\text{norm}}(f_{\text{trans}}(W))$, where $f_{\text{trans}}$ and $f_{\text{norm}}$ are a normalization and a transformation operators, respectively.

The operator $f_{\text{trans}}$ can be realized with each of the three following functions: 

\begin{enumerate}
    \item \textbf{Absolute Value Transformation}: 
    For each element \( W_{i,j} \) of the weight matrix \( W \), we apply the absolute value function:
    \[
    f_{\text{trans}}^{\text{abs}}(W_{i,j}) = |W_{i,j}|
    \]

    \item \textbf{Logarithmic Transformation (Log with Offset)}: 
    First, we take the absolute value of the weights, then add a constant (usually 1) to avoid taking the logarithm of zero, and finally apply the logarithm. This transformation is given by:
    \[
    f_{\text{trans}}^{\text{log}}(W_{i,j}) = \log(|W_{i,j}| + 1)
    \]


    \item \textbf{Square Transformation}: 
    Each element \( W_{i,j} \) of the weight matrix is squared to ensure positivity and a more concentrated mapping:
    \[
    f_{\text{trans}}^{\text{square}}(W_{i,j}) = W_{i,j}^2
    \]
    
\end{enumerate}

We default set $f_{\text{trans}}$ to the logarithmic transformation $f_{\text{trans}}^{\text{log}}$ because the derivative of the log function is an inverse function, meaning that larger values of the weights result in smaller gradients during optimization. This leads to a reduction in the rate at which the weights grow, making it more difficult for the weights to reach excessively large values.  
We also evaluate the other two kinds of transformations in Section \ref{}.
The experimental results prove that Section \label{}. 

After applying any of these transformations to the weights, we deploy a normalization to the second dimension (the dimension corresponding to the channels) of the weight to ensure that the weights sum to one along this dimension. Specifically, we normalize the weights by dividing each column of the transformed matrix by the sum of the elements in that column.
Therefore, we can formulate $f_{\text{sim}}$ as
\[
f_{\text{\text{sim}}}(W_{i,j}) = f_{\text{norm}} (f_{\text{trans}}(W_{i,j})) = \frac{f_{\text{trans}}(W_{i,j})}{\sum_{j=1}^N f_{\text{trans}}(W_{i,j})}
\]

Finally, the normalized weight matrix \( f_{\text{sim}}(W_{i,j}) \) is guaranteed to lie on the Standard N-Simplex, as the weights are positive, sum to one, and respect the structural constraints of the simplex.
We provide theoretical analysis in Section \ref{sec:proof}.




\subsection{Overall Architecture}
Our FSMLP is composed of two kinds of blocks: Frequency \textbf{S}implex \textbf{C}hannel-\textbf{W}ise \textbf{M}LP (\textbf{SCWM}) and \textbf{F}requency \textbf{T}emporal \textbf{M}LP (\textbf{FTM}).
Fig. \ref{fig:model} shows the architecture of the proposed FSMLP, where $N$ SCWM blocks and $N$ FTM blocks are cascaded respectively to gradually capture the complex inter-channel dependencies and frequency temporal relations.
As mentioned above, each frequency component corresponds to a specific period in the time domain. 
By modeling inter-channel dependencies in the frequency domain, the model captures variations across different periods in different channels, which helps mitigate overfitting to noise in time series data, as it focuses on the underlying periodic patterns rather than the noise inherent in the time-domain representation \cite{frets,fedformer}.
We perform Frequency Transformation (FT) to convert the input into the frequency domain, allowing the model to process frequency representations rather than time-domain data.
After the linear forecasting head, we apply an inverse Frequency Transformation (iFT) to convert the frequency domain output back to the time domain for the final prediction.




\subsubsection{SCWM Block}
The SCWM block consists of two main steps. 
First, to extract inter-channel information, we use the proposed Simplex MLP to capture inter-channel dependencies. For the $l$-th SCWM block, we formulate this step as follows:
\[
\mathbf{Z}^l_{\text{Channel}} = \sigma(f_\text{sim}(\mathbf{Z}^{l-1}_{\text{SCWM}})) +\mathbf{Z}^{l-1}_{\text{SCWM}}
\]
where \(\mathbf{Z}^{l-1}_{\text{SCWM}}\) represents the output from the $l$-1-th SCWM block, and \(\sigma\) is the activation function applied to the output.

Next, we extract the temporal information using a simple one-layer MLP as follows:

\[
\mathbf{Z}^{l}_{\text{SCWM}} = \sigma(\text{MLP}(\mathbf{Z}^l_{\text{Channel}})) +\mathbf{Z}^l_{\text{Channel}}
\]

where \(\mathbf{Z}^{l}_{\text{SCWM}}\) represents the output of the \(l\)-th block SCWM.

\subsubsection{FTM Block}
Given the output of the $N$-th SCWM block: $\mathbf{Z}^{N}_{\text{SCWM}}$, we apply $N$ cascaded FTM blocks to further capture temporal dependencies in the time series.
This process can be expressed as:
\[
\mathbf{Z}_\text{FTM}^{i} = \sigma(Linear(\mathbf{Z}_\text{FTM}^{i-1}))+\mathbf{Z}_{\text{FTM}}^{i-1},
\]
where \(\mathbf{Z}_{\text{FTM}}^{i-1}\) represents the output from the $i-1$ th FTM block.
Then, we utilize a linear layer as our forecast head, as:
\[
\mathbf{\hat{Y}}=Linear(\mathbf{Z}_\text{FTM}^N),
\]
where $\mathbf{\hat{Y}}$ is the prediction.

\subsection{Loss Function}
To thoroughly leverage the advantages of the frequency and time domains, we propose calculating the loss function specifically in each domain. 
For the time domain, we use Mean Squared Error (MSE) as the loss function, while for the frequency domain, we utilize Mean Absolute Error (MAE) loss instead of MSE.
The reason why we use L1 loss function in frequency domain is that different frequency components often exhibit vastly varying magnitudes, rendering squared loss methods unstable. 
The overall loss function of our method can be expressed as:
\begin{equation}
    \left\{
    \begin{array}{lr}
    \mathcal{L}_\text{time} =  \sum_{i=1}^\tau\frac{||\mathbf{Y}_i - F(\mathbf{X})_i||^2}{\tau}, \\
    \mathcal{L}_\text{fre} =  \sum_{i=1}^\tau\frac{||\mathbf{Y}_i - F(\mathbf{X})_i||}{\tau}, \\
    \mathcal{L}_\text{total} = \mathcal{L}_\text{time} + \mathcal{L}_\text{fre}.
    \end{array}
    \right.
\end{equation}

\vspace{10pt}

\section{Theoretical Analysis}
\label{sec:proof}
In this section, we compare the Rademacher complexity \cite{rademacher} of standard MLP and our proposed Simplex-MLP, specifically in the context of regression tasks. This comparison will shed light on why the Simplex-MLP, with its weight constraints, is less prone to overfitting compared to the standard MLP, particularly in the presence of extreme values or noisy data.

For a standard MLP, the upper bound of  Rademacher complexity, which measures the model's capacity to fit random noise, can be expressed as follows:

\[
\mathcal{R}_S(\mathcal{H}) \leq \frac{B}{m} \sqrt{\sum_{i=1}^m \| x^{(i)} \|_2^2},
\]
where \( \mathcal{H} \) is the hypothesis class of the MLP. Here, \( w \in \mathbb{R}^d \) are the weight parameters of the model, with \( d \) representing the dimensionality of the input, and \( m \) is the number of data points in the training set. The term \( x^{(i)} \in \mathbb{R}^d \) denotes the \( i \)-th input data point, and \( \| x^{(i)} \|_2 \) is the \( \ell_2 \)-norm. The constant \( B \) is an upper bound on the norm of the weight vector \( w \), often assumed to be \( B = \| w \|_2 \).

This bound indicates that the capacity of a standard MLP is influenced by both the norm of the weight vector \( B \) and the sum of the squared \( \ell_2 \)-norms of the input data points. Given that the weight parameters are unconstrained, the model possesses significant flexibility to adapt to the data, including any outliers. Consequently, this adaptability can elevate the risk of overfitting, particularly in the presence of extreme values or noise within the data.

Now, consider the Simplex-MLP, where the weight vector is constrained to lie within the standard \( n \)-simplex. The standard \( n \)-simplex \( \Delta^n \) is defined as:

\[
\Delta^n = \left\{ w \in \mathbb{R}^n \mid w_i \geq 0 \text{ for all } i, \sum_{i=1}^n w_i = 1 \right\}
\]

This constraint on the weights ensures that each component \( w_i \) is non-negative and that the sum of the weights is exactly $1$. This constraint enforces that the model cannot assign disproportionately large weights to any feature, preventing it from overfitting to noisy or extreme values in the data.

The upper bound of Rademacher complexity of the Simplex-MLP is given by:

\[
\mathcal{R}_S(\mathcal{H}_{\Delta}) \leq \frac{1}{m} \sqrt{\sum_{i=1}^m \| x^{(i)} \|_2^2},
\]
where \( \mathcal{H}_{\Delta} \) represents the hypothesis class of the Simplex-MLP, in which the weights are constrained to lie within the standard \( n \)-simplex \( \Delta^n \). 
The key distinction here is that the weight vector \( w \in \Delta^n \) must satisfy \( w_i \geq 0 \) and \( \sum_{i=1}^n w_i = 1 \). This constraint significantly limits the model's flexibility in assigning disproportionately large weights to any specific feature, thereby potentially reducing the risk of overfitting.




\begin{theorem}
    Let \( \Delta^n \) denote the standard \( n \)-simplex: $
\Delta^n = \left\{ w \in \mathbb{R}^n \mid w_i \geq 0, \sum_{i=1}^n w_i = 1 \right\}$, \( \mathcal{H}_{\Delta} \)  be the hypothesis class with weights in the \( n \)-simplex:
$
\mathcal{H}_{\Delta} = \left\{ f_w(x) = w^\top x \mid w \in \Delta^n \right\},
$
where \( f_w(x) \) is a linear function of the data point \( x \) with weight vector \( w \in \Delta^n \).
The Rademacher complexity \( \mathcal{R}_S(\mathcal{H}_{\Delta}) \) of the hypothesis class \( \mathcal{H}_{\Delta} \) is bounded by:
\[
\mathcal{R}_S(\mathcal{H}_{\Delta}) \leq \frac{1}{m} \sqrt{\sum_{i=1}^m \| x^{(i)} \|_2^2},
\]
where \( x^{(i)} \) are the data points.
\end{theorem}

\begin{proof}
    The Rademacher complexity \( \mathcal{R}_S(\mathcal{H}_{\Delta}) \) is given by:

    \begin{equation*}
        \begin{aligned}
            \mathcal{R}_S(\mathcal{H}_{\Delta}) 
            &= \mathbb{E}_\sigma \left[ \sup_{w \in \Delta^n} \frac{1}{m} \sum_{i=1}^m \sigma_i \left\langle w, x^{(i)} \right\rangle \right]\\
            &= \frac{1}{m} \mathbb{E}_\sigma \left[ \sup_{w \in \Delta^n} \left\langle w, \sum_{i=1}^m \sigma_i x^{(i)} \right\rangle \right] \\
            &= \frac{1}{m} \mathbb{E}_\sigma \left[ \sup_{w \in \Delta^n} w^\top v \right],
        \end{aligned}
    \end{equation*}
where we denote \( v = \sum_{i=1}^n \sigma_i x^{(i)} \). 
Since \( w \in \Delta^n \), the supremum is attained when \( w \) is aligned with \( v \), that is:
\[
\sup_{w \in \Delta^n} w^\top v = \| v \|_2
\]

Thus, the Rademacher complexity becomes:
\[
\mathcal{R}_S(\mathcal{H}_{\Delta}) = \frac{1}{m} \mathbb{E}_\sigma \left[ \| v \|_2 \right]\leq \frac{1}{m} \sqrt{\mathbb{E}_\sigma \left[ \| v \|_2^2 \right]}
\]
which immediately follows from the Jensen's inequality for the convex function \( \|\cdot\|_2 \).
Furthermore,  expand \( v \) and compute \( \mathbb{E}_\sigma \left[ \| v \|_2^2 \right] \):
\[
\mathbb{E}_\sigma \left[ \| v \|_2^2 \right] = \mathbb{E}_\sigma \left[ \sum_{i=1}^m \| \sigma_i x^{(i)} \|_2^2 + \sum_{i=1}^m\sum_{j \neq i} \langle \sigma_i x^{(i)}, \sigma_j x^{(j)} \rangle \right]
\]

Since \( \sigma_i \) are independent and \( \mathbb{E}[\sigma_i^2] = 1 \), the cross terms vanish, and we are left with:
\[
\mathbb{E}_\sigma \left[ \| v \|_2^2 \right] = \sum_{i=1}^n \| x^{(i)} \|_2^2
\]

Thus, the Rademacher complexity is bounded by:
\[
\mathcal{R}_S(\mathcal{H}_{\Delta}) \leq \frac{1}{m} \sqrt{\sum_{i=1}^m \| x^{(i)} \|_2^2}
\]
\end{proof}

\vspace{10pt}
\section{Experiments}

\begin{table}
   \caption{Dataset description.}
   \label{tab:datasets}
   \footnotesize
   \renewcommand{\arraystretch}{1.2}
   \setlength{\tabcolsep}{2pt}
   \centering
      \begin{tabular}{c|cllcccc}
        \toprule
         {Datasets}    & {ETTh1}& {ETTh2}&{ETTm1}& {ETTm2}& {Traffic} & {ECL} & {Weather}      \\ 
         \midrule
         Channels    & 7         & 7         
&7        
& 7        & 862     & 321         & 21      \\
         Timesteps   & 17,420    & 17,420    
&69,680   
& 69,680   & 17,544  & 26,304      & 52,696  \\
         Granularity & 1 hour    & 1 hour    &5 min    & 5 min    & 1 hour  & 1 hour      & 10 min \\
         \bottomrule
      \end{tabular}
\end{table}

\begin{table*}
  \caption{Full results on the long-term forecasting task with forecast lengths $\tau=96, 192, 336 \text{ and } 720$. The length of history window is set to 96 for all baselines. \emph{Avg} indicates the results averaged over forecasting lengths.}\label{tab:main1}
  \vskip 0.1in
  \renewcommand{\arraystretch}{1.1} 
  \setlength{\tabcolsep}{1.5pt}
  \centering
  \small
  \renewcommand{\multirowsetup}{\centering}
  \begin{tabular}{c|c|cc|cc|cc|cc|cc|cc|cc|cc|cc|cc|cc|cc}
    \toprule
    \multicolumn{2}{l}{\multirow{2}{*}{\rotatebox{0}{\scaleb{Models}}}} & 
    \multicolumn{2}{c}{\rotatebox{0}{\scaleb{\textbf{FSMLP}}}} &
    \multicolumn{2}{c}{\rotatebox{0}{\scaleb{{iTransformer}}}} &
    \multicolumn{2}{c}{\rotatebox{0}{\scaleb{{FreTS}}}} &
    \multicolumn{2}{c}{\rotatebox{0}{\scaleb{{TSMixer}}}} &
    \multicolumn{2}{c}{\rotatebox{0}{\scaleb{TimesNet}}} &
    \multicolumn{2}{c}{\rotatebox{0}{\scaleb{Crossformer}}}  &
    \multicolumn{2}{c}{\rotatebox{0}{\scaleb{TiDE}}} &
    \multicolumn{2}{c}{\rotatebox{0}{\scaleb{{DLinear}}}} &
    \multicolumn{2}{c}{\rotatebox{0}{\scaleb{FEDformer}}} &
    \multicolumn{2}{c}{\rotatebox{0}{\scaleb{PatchTST}}} &
    \multicolumn{2}{c}{\rotatebox{0}{\scaleb{Autoformer}}} &
    \multicolumn{2}{c}{\rotatebox{0}{\scaleb{FITS}}} \\
    \multicolumn{2}{c}{} &
    \multicolumn{2}{c}{\scaleb{\textbf{(Ours)}}} & 
    \multicolumn{2}{c}{\scaleb{(2024)}} &
    \multicolumn{2}{c}{\scaleb{(2023)}} &
    \multicolumn{2}{c}{\scaleb{(2023)}} & 
    \multicolumn{2}{c}{\scaleb{(2023)}} & 
    \multicolumn{2}{c}{\scaleb{(2023)}} & 
    \multicolumn{2}{c}{\scaleb{(2023)}} & 
    \multicolumn{2}{c}{\scaleb{(2023)}} &
    \multicolumn{2}{c}{\scaleb{(2022)}} &
    \multicolumn{2}{c}{\scaleb{(2023)}} &
    \multicolumn{2}{c}{\scaleb{(2021)}} &
    \multicolumn{2}{c}{\scaleb{(2024)}} \\
    \cmidrule(lr){3-4} \cmidrule(lr){5-6}\cmidrule(lr){7-8} \cmidrule(lr){9-10}\cmidrule(lr){11-12}\cmidrule(lr){13-14} \cmidrule(lr){15-16} \cmidrule(lr){17-18} \cmidrule(lr){19-20} \cmidrule(lr){21-22} \cmidrule(lr){23-24} \cmidrule(lr){25-26}
    \multicolumn{2}{l}{\rotatebox{0}{\scaleb{Metrics}}}  & \scalea{MSE} & \scalea{MAE}  & \scalea{MSE} & \scalea{MAE}  & \scalea{MSE} & \scalea{MAE}  & \scalea{MSE} & \scalea{MAE}  & \scalea{MSE} & \scalea{MAE}  & \scalea{MSE} & \scalea{MAE} & \scalea{MSE} & \scalea{MAE} & \scalea{MSE} & \scalea{MAE} & \scalea{MSE} & \scalea{MAE} & \scalea{MSE} & \scalea{MAE}  & \scalea{MSE} & \scalea{MAE} & \scalea{MSE} & \scalea{MAE}  \\
    \toprule
    
    \multirow{5}{*}{{\rotatebox{90}{\scalebox{0.95}{ETTm1}}}}
    & \scalea{96} & \scalea{\bst{0.303}} & \scalea{\bst{0.342}} & \scalea{0.334} & \scalea{0.368} & \scalea{0.339} & \scalea{{0.374}} & \scalea{{0.479}} & \scalea{{0.470}}& 
    \scalea{{0.338}} & \scalea{0.375} & \scalea{0.375} & \scalea{0.415} & \scalea{0.364} & \scalea{0.387} & \scalea{0.345} & \scalea{0.372} & \scalea{0.389} & \scalea{0.427} & \scalea{\subbst{0.329}} & \scalea{\subbst{0.367}} & \scalea{0.468} & \scalea{0.463} & \scalea{0.365} & \scalea{0.380} \\
    & \scalea{192} & \scalea{\bst{0.347}} & \scalea{\bst{0.368}} & \scalea{0.377} & \scalea{0.391} & \scalea{0.382} & \scalea{0.397} & \scalea{{0.480}} & \scalea{{0.482}}& \scalea{0.374} & \scalea{0.387} & \scalea{0.453} & \scalea{0.474} & \scalea{0.398} & \scalea{0.404} & \scalea{{0.381}} & \scalea{{0.390}} & \scalea{0.402} & \scalea{0.431}  & \scalea{\subbst{0.367}} & \scalea{\subbst{0.385}} & \scalea{0.573} & \scalea{0.509} & \scalea{0.400} & \scalea{0.397}  \\
    & \scalea{336} & \scalea{\bst{0.378}} & \scalea{\bst{0.391}} & \scalea{0.426} & \scalea{0.420} & \scalea{0.421} & \scalea{0.426} & \scalea{{0.541}} & \scalea{{0.525}}& \scalea{0.410} & \scalea{0.411} & \scalea{0.548} & \scalea{0.526} & \scalea{0.428} & \scalea{0.425} & \scalea{{0.414}} & \scalea{{0.414}} & \scalea{0.438} & \scalea{0.451} & \scalea{\subbst{0.399}} & \scalea{\subbst{0.410}} & \scalea{0.596} & \scalea{0.527} & \scalea{0.431} & \scalea{0.418} \\
    & \scalea{720} & \scalea{\bst{0.433}} & \scalea{\bst{0.428}} & \scalea{0.491} & \scalea{0.459} & \scalea{0.485} & \scalea{0.462} & \scalea{{0.616}} & \scalea{{0.574}}& \scalea{0.478} & \scalea{0.450} & \scalea{0.857} & \scalea{0.713} & \scalea{0.487} & \scalea{0.461} & \scalea{{0.473}} & \scalea{{0.451}} & \scalea{0.529} & \scalea{0.498} & \scalea{\subbst{0.454}} & \scalea{\subbst{0.439}} & \scalea{0.749} & \scalea{0.569} & \scalea{0.492} & \scalea{0.491} \\
    \cmidrule(lr){2-26}
    & \scalea{Avg} & \scalea{\bst{0.365}} & \scalea{\bst{0.382}} & \scalea{0.407} & \scalea{0.410} & \scalea{0.407} & \scalea{0.415} & \scalea{0.529} & \scalea{0.513}& \scalea{0.400} & \scalea{0.406} & \scalea{0.558} & \scalea{0.532} & \scalea{0.419} & \scalea{0.419} & \scalea{{0.404}} & \scalea{{0.407}} & \scalea{0.440} & \scalea{0.451}  & \scalea{\subbst{0.387}} & \scalea{\subbst{0.400}} & \scalea{0.596} & \scalea{0.517} & \scalea{0.422} & \scalea{0.421}  \\
    \midrule
    
    \multirow{5}{*}{{\rotatebox{90}{\scalebox{0.95}{ETTm2}}}}
    & \scalea{96}  & \scalea{\bst{0.166}} & \scalea{\bst{0.247}} & \scalea{{0.180}} & \scalea{0.264} & \scalea{0.190} & \scalea{0.282}  & \scalea{0.250} & \scalea{0.366}&\scalea{0.185} & \scalea{{0.264}} & \scalea{0.267} & \scalea{0.349} & \scalea{0.207} & \scalea{0.305} & \scalea{0.195} & \scalea{0.294} & \scalea{0.194} & \scalea{0.284} & \scalea{\subbst{0.175}} & \scalea{\subbst{0.259}} & \scalea{0.240} & \scalea{0.319} & \scalea{0.186} & \scalea{0.269}  \\
    & \scalea{192} & \scalea{\bst{0.229}} & \scalea{\bst{0.289}} & \scalea{0.250} & \scalea{0.309} & \scalea{0.260} & \scalea{0.329} & \scalea{0.492} & \scalea{0.559} & \scalea{{0.254}} & \scalea{{0.307}} & \scalea{0.472} & \scalea{0.479} & \scalea{0.290} & \scalea{0.364} & \scalea{0.283} & \scalea{0.359} & \scalea{0.264} & \scalea{0.324}  & \scalea{\subbst{0.241}} & \scalea{\subbst{0.302}} & \scalea{0.300} & \scalea{0.349} & \scalea{0.249} & \scalea{0.306}  \\
    & \scalea{336} & \scalea{\bst{0.286}} & \scalea{\bst{0.326}} & \scalea{0.311} & \scalea{0.348} & \scalea{0.373} & \scalea{0.405}  & \scalea{0.833} & \scalea{0.734}& \scalea{{0.314}} & \scalea{{0.345}} & \scalea{0.919} & \scalea{0.702} & \scalea{0.377} & \scalea{0.422} & \scalea{0.384} & \scalea{0.427} & \scalea{0.319} & \scalea{0.359}  & \scalea{\subbst{0.305}} & \scalea{0.343} & \scalea{0.339} & \scalea{0.375} & \scalea{0.309} & \scalea{\subbst{0.343}} \\
    & \scalea{720} & \scalea{\bst{0.380}} & \scalea{\bst{0.383}} & \scalea{{0.412}} & \scalea{{0.407}} & \scalea{0.517} & \scalea{0.499}  & \scalea{2.543} & \scalea{1.352}& \scalea{0.408} & \scalea{0.403} & \scalea{4.874} & \scalea{1.601} & \scalea{0.558} & \scalea{0.524} & \scalea{0.516} & \scalea{0.502} & \scalea{0.430} & \scalea{0.424} & \scalea{\subbst{0.402}} & \scalea{{0.400}} & \scalea{0.423} & \scalea{0.421} & \scalea{0.410} & \scalea{\subbst{0.398}}  \\
    \cmidrule(lr){2-26}
    & \scalea{Avg} & \scalea{\bst{0.265}} & \scalea{\bst{0.311}} & \scalea{{0.288}} & \scalea{0.332} & \scalea{0.335} & \scalea{0.379}  & \scalea{1.030} & \scalea{0.753}& \scalea{0.297} & \scalea{{0.329}} & \scalea{1.633} & \scalea{0.782} & \scalea{0.358} & \scalea{0.404} & \scalea{0.344} & \scalea{0.396} & \scalea{0.302} & \scalea{0.348}  & \scalea{\subbst{0.281}} & \scalea{\subbst{0.347}} & \scalea{0.326} & \scalea{0.366} & \scalea{0.289} & \scalea{0.351} \\
    \midrule
    
    \multirow{5}{*}{\rotatebox{90}{{\scalebox{0.95}{ETTh1}}}}
    & \scalea{96} & \scalea{\bst{0.361}} & \scalea{\bst{0.384}} & \scalea{0.386} & \scalea{{0.405}} & \scalea{0.399} & \scalea{0.412}  & \scalea{0.466} & \scalea{0.482}& \scalea{0.384} & \scalea{0.402} & \scalea{0.441} & \scalea{0.457} & \scalea{0.479} & \scalea{0.464} & \scalea{0.396} & \scalea{0.410} & \scalea{\subbst{0.377}} & \scalea{0.418}  & \scalea{0.414} & \scalea{0.419} & \scalea{0.423} & \scalea{0.441} & \scalea{0.387} & \scalea{\subbst{0.394}}  \\
    & \scalea{192} & \scalea{\bst{0.405}} & \scalea{\bst{0.419}} & \scalea{0.441} & \scalea{{0.436}} & \scalea{0.453} & \scalea{0.443}  & \scalea{0.597} & \scalea{0.567}& \scalea{0.436} & \scalea{0.429} & \scalea{0.521} & \scalea{0.503} & \scalea{0.525} & \scalea{0.492} & \scalea{0.449} & \scalea{0.444} & \scalea{\subbst{0.421}} & \scalea{0.445}  & \scalea{0.460} & \scalea{0.445} & \scalea{0.498} & \scalea{0.485} & \scalea{0.436} & \scalea{\subbst{0.422}} \\
    & \scalea{336} & \scalea{\bst{0.444}} & \scalea{\bst{0.440}} & \scalea{0.487} & \scalea{{0.458}} & \scalea{0.503} & \scalea{0.475}  & \scalea{0.677} & \scalea{0.618}& \scalea{0.491} & \scalea{0.469} & \scalea{0.659} & \scalea{0.603} & \scalea{0.565} & \scalea{0.515} & \scalea{0.487} & \scalea{0.465} & \scalea{\subbst{0.468}} & \scalea{0.466}  & \scalea{0.501} & \scalea{0.463} & \scalea{0.506} & \scalea{0.496} & \scalea{0.478} & \scalea{\subbst{0.443}} \\
    & \scalea{720} & \scalea{\bst{0.454}} & \scalea{\bst{0.457}} & \scalea{0.503} & \scalea{{0.491}} & \scalea{0.596} & \scalea{0.565}  & \scalea{0.752} & \scalea{0.674}& \scalea{0.521} & \scalea{0.500} & \scalea{0.893} & \scalea{0.736} & \scalea{0.594} & \scalea{0.558} & \scalea{0.516} & \scalea{0.513} & \scalea{0.500} & \scalea{0.493}  & \scalea{0.500} & \scalea{0.488} & \scalea{{0.477}} & \scalea{0.487} & \scalea{\subbst{0.468}} & \scalea{\subbst{0.463}} \\
    \cmidrule(lr){2-26}
    & \scalea{Avg} & \scalea{\bst{0.416}} & \scalea{\bst{0.425}} & \scalea{0.454} & \scalea{{0.497}} & \scalea{0.488} & \scalea{0.474}  & \scalea{0.623} & \scalea{0.585}& \scalea{0.458} & \scalea{0.450} & \scalea{0.628} & \scalea{0.574} & \scalea{0.541} & \scalea{0.507} & \scalea{0.462} & \scalea{0.458} & \scalea{\subbst{0.441}} & \scalea{0.457}  & \scalea{0.469} & \scalea{0.454} & \scalea{0.476} & \scalea{0.477} & \scalea{{0.442}} & \scalea{\subbst{0.430}}  \\
    \midrule

    \multirow{5}{*}{\rotatebox{90}{\scalebox{0.95}{ETTh2}}}
    & \scalea{96}  & \scalea{\bst{0.277}} & \scalea{\bst{0.328}} & \scalea{{0.297}} & \scalea{{0.349}} & \scalea{0.350} & \scalea{0.403}  & \scalea{1.056} & \scalea{0.806}& \scalea{0.320} & \scalea{0.364} & \scalea{0.681} & \scalea{0.592} & \scalea{0.400} & \scalea{0.440} & \scalea{0.343} & \scalea{0.396} & \scalea{0.347} & \scalea{0.391}  & \scalea{0.302} & \scalea{0.348} & \scalea{0.383} & \scalea{0.424} & \scalea{\subbst{0.293}} & \scalea{\subbst{0.340}} \\
    & \scalea{192} & \scalea{\bst{0.346}} & \scalea{\bst{0.377}} & \scalea{{0.380}} & \scalea{{0.400}} & \scalea{0.472} & \scalea{0.475}  & \scalea{2.586} & \scalea{1.403}& \scalea{0.409} & \scalea{0.417} & \scalea{1.837} & \scalea{1.054} & \scalea{0.528} & \scalea{0.509} & \scalea{0.473} & \scalea{0.474} & \scalea{0.430} & \scalea{0.443}  & \scalea{0.388} & \scalea{0.400} & \scalea{0.557} & \scalea{0.511} & \scalea{\subbst{0.378}} & \scalea{\subbst{0.391}}  \\
    & \scalea{336} & \scalea{\bst{0.385}} & \scalea{\bst{0.408}} & \scalea{{0.428}} & \scalea{{0.432}} & \scalea{0.564} & \scalea{0.528}  & \scalea{2.407} & \scalea{1.348}& \scalea{0.449} & \scalea{0.451} & \scalea{3.000} & \scalea{1.472} & \scalea{0.643} & \scalea{0.571} & \scalea{0.603} & \scalea{0.546} & \scalea{0.469} & \scalea{0.475}  & \scalea{0.426} & \scalea{0.433}& \scalea{0.470} & \scalea{0.481} & \scalea{\subbst{0.418}} & \scalea{\subbst{0.425}}  \\
    & \scalea{720} & \scalea{\bst{0.394}} & \scalea{\bst{0.424}} & \scalea{{0.427}} & \scalea{{0.445}} & \scalea{0.815} & \scalea{0.654} & \scalea{2.051} & \scalea{1.218} & \scalea{0.473} & \scalea{0.474} & \scalea{3.024} & \scalea{1.399} & \scalea{0.874} & \scalea{0.679} & \scalea{0.812} & \scalea{0.650} & \scalea{0.473} & \scalea{0.480}  & \scalea{0.431} & \scalea{0.446} & \scalea{0.501} & \scalea{0.515} & \scalea{\subbst{0.419}} & \scalea{\subbst{0.436}}  \\
    \cmidrule(lr){2-26}
    & \scalea{Avg} & \scalea{\bst{0.350}} & \scalea{\bst{0.384}} & \scalea{{0.383}} & \scalea{{0.407}} & \scalea{0.550} & \scalea{0.515}  & \scalea{2.025} & \scalea{1.194}& \scalea{0.413} & \scalea{0.426} & \scalea{2.136} & \scalea{1.130} & \scalea{0.611} & \scalea{0.550} & \scalea{0.558} & \scalea{0.516} & \scalea{0.430} & \scalea{0.447}  & \scalea{0.387} & \scalea{0.407} & \scalea{0.478} & \scalea{0.483} & \scalea{\subbst{0.377}} & \scalea{\subbst{0.398}}  \\
    \midrule
    
    \multirow{5}{*}{\rotatebox{90}{\scalebox{0.95}{ECL}}} 
    & \scalea{96} & \scalea{\bst{0.133}} & \scalea{\bst{0.226}} & \scalea{\subbst{0.148}} & \scalea{\subbst{0.239}} & \scalea{0.189} & \scalea{0.277}  & \scalea{0.204} & \scalea{0.308}& \scalea{0.171} & \scalea{0.273} & \scalea{\subbst{0.148}} & \scalea{0.248} & \scalea{0.237} & \scalea{0.329} & \scalea{0.210} & \scalea{0.302} & \scalea{0.200} & \scalea{0.315}  & \scalea{0.181} & \scalea{0.270} & \scalea{0.199} & \scalea{0.315} & \scalea{0.205} & \scalea{0.280} \\
    & \scalea{192} & \scalea{\bst{0.150}} & \scalea{\bst{0.242}} & \scalea{0.162} & \scalea{\subbst{0.253}} & \scalea{0.193} & \scalea{0.282}  & \scalea{0.218} & \scalea{0.329}& \scalea{0.188} & \scalea{0.289} & \scalea{\subbst{0.161}} & \scalea{0.263} & \scalea{0.236} & \scalea{0.330} & \scalea{0.210} & \scalea{0.305} & \scalea{0.207} & \scalea{0.322}  & \scalea{0.188} & \scalea{0.274} & \scalea{0.215} & \scalea{0.327} & \scalea{0.202} & \scalea{0.281}  \\
    & \scalea{336} & \scalea{\bst{0.164}} & \scalea{\bst{0.259}} & \scalea{\subbst{0.178}} & \scalea{\subbst{0.269}} & \scalea{0.207} & \scalea{0.296}  & \scalea{0.239} & \scalea{0.350}& \scalea{0.208} & \scalea{0.304} & \scalea{0.191} & \scalea{0.289} & \scalea{0.249} & \scalea{0.344} & \scalea{0.223} & \scalea{0.319} & \scalea{0.226} & \scalea{0.340}  & \scalea{0.204} & \scalea{0.293} & \scalea{0.232} & \scalea{0.343} & \scalea{0.217} & \scalea{0.297} \\
    & \scalea{720} & \scalea{\bst{0.187}} & \scalea{\bst{0.280}} & \scalea{\subbst{0.225}} & \scalea{\subbst{0.317}} & \scalea{0.245} & \scalea{0.332}  & \scalea{0.272} & \scalea{0.373}& \scalea{0.289} & \scalea{0.363} & \scalea{0.226} & \scalea{0.314} & \scalea{0.284} & \scalea{0.373} & \scalea{0.258} & \scalea{0.350} & \scalea{0.282} & \scalea{0.379}  & \scalea{0.246} & \scalea{0.324} & \scalea{0.268} & \scalea{0.371} & \scalea{0.261} & \scalea{0.332}\\
    \cmidrule(lr){2-26}
    & \scalea{Avg} & \scalea{\bst{0.159}} & \scalea{\bst{0.252}} & \scalea{\subbst{0.178}} & \scalea{\subbst{0.270}} & \scalea{0.209} & \scalea{0.297}  & \scalea{0.233} & \scalea{0.340}& \scalea{0.214} & \scalea{0.307} & \scalea{0.182} & \scalea{0.279} & \scalea{0.251} & \scalea{0.344} & \scalea{0.225} & \scalea{0.319} & \scalea{0.229} & \scalea{0.339}  & \scalea{0.205} & \scalea{0.290} & \scalea{0.228} & \scalea{0.339} & \scalea{0.224} & \scalea{0.298}  \\
    \midrule
    
    \multirow{5}{*}{\rotatebox{90}{\scalebox{0.95}{Traffic}}} 
    & \scalea{96} & \scalea{\bst{0.379}} & \scalea{\bst{0.254}}  & \scalea{\subbst{0.395}} & \scalea{\subbst{0.268}} & \scalea{0.528} & \scalea{0.341} & \scalea{0.531} & \scalea{0.358} & \scalea{0.518} & \scalea{0.269} & \scalea{0.805} & \scalea{0.493} & \scalea{0.697} & \scalea{0.429} & \scalea{0.577} & \scalea{0.362} & \scalea{0.609} & \scalea{0.385} & \scalea{0.462} & \scalea{0.295} & \scalea{1.451} & \scalea{0.744} & \scalea{0.686} & \scalea{0.405} \\
    & \scalea{192} & \scalea{\bst{0.405}} & \scalea{\bst{0.264}} & \scalea{\subbst{0.417}} & \scalea{\subbst{0.276}} & \scalea{0.531} & \scalea{0.338} & \scalea{0.566} & \scalea{0.387} & \scalea{0.551} & \scalea{0.285} & \scalea{0.756} & \scalea{0.474} & \scalea{0.647} & \scalea{0.407} & \scalea{0.603} & \scalea{0.372} & \scalea{0.633} & \scalea{0.400} & \scalea{0.466} & \scalea{0.296} & \scalea{0.842} & \scalea{0.622} & \scalea{0.623} & \scalea{0.374} \\
    & \scalea{336} & \bst{\scalea{0.422}} & \bst{\scalea{0.274}} & \scalea{\subbst{0.432}} & \scalea{\subbst{0.283}} & \scalea{0.551} & \scalea{0.345} & \scalea{0.578} & \scalea{0.392} & \scalea{0.546} & \scalea{0.293} & \scalea{0.762} & \scalea{0.477} & \scalea{0.653} & \scalea{0.410} & \scalea{0.615} & \scalea{0.378} & \scalea{0.637} & \scalea{0.398} & \scalea{0.482} & \scalea{0.304} & \scalea{0.844} & \scalea{0.620} & \scalea{0.629} & \scalea{0.378} \\
    & \scalea{720} & \bst{\scalea{0.453}} & \bst{\scalea{0.294}} & \scalea{\subbst{0.467}} & \scalea{\subbst{0.302}} & \scalea{0.598} & \scalea{0.367} & \scalea{0.617} & \scalea{0.415} & \scalea{0.597} & \scalea{0.323} & \scalea{0.719} & \scalea{0.449} & \scalea{0.694} & \scalea{0.429} & \scalea{0.649} & \scalea{0.403} & \scalea{0.668} & \scalea{0.415} & \scalea{0.514} & \scalea{0.322} & \scalea{0.867} & \scalea{0.624} & \scalea{0.668} & \scalea{0.396} \\
    \cmidrule(lr){2-26}
    & \scalea{Avg} & \bst{\scalea{0.415}} & \bst{\scalea{0.272}} & \scalea{{\subbst{0.428}}} & \scalea{\subbst{0.282}} & \scalea{0.552} & \scalea{0.348} & \scalea{0.573} & \scalea{0.388} & \scalea{0.553} & \scalea{0.292} & \scalea{0.760} & \scalea{0.473} & \scalea{0.673} & \scalea{0.419} & \scalea{0.611} & \scalea{0.379} & \scalea{0.637} & \scalea{0.399} & \scalea{0.481} & \scalea{0.304} & \scalea{1.001} & \scalea{0.652} & \scalea{0.652} & \scalea{0.388} \\
    \midrule
    
    \multirow{5}{*}{\rotatebox{90}{\scalebox{0.95}{Weather}}} 
    & \scalea{96} & \scalea{\bst{0.149}} & \scalea{\bst{0.193}} & \scalea{0.174} & \scalea{0.214} & \scalea{0.184} & \scalea{0.239}  & \scalea{0.180} & \scalea{0.252}& \scalea{0.178} & \scalea{{0.226}} & \scalea{{0.177}} & \scalea{0.246} & \scalea{0.202} & \scalea{0.261} & \scalea{0.197} & \scalea{0.259} & \scalea{0.221} & \scalea{0.304}  & \scalea{0.177} & \scalea{0.218}& \scalea{0.284} & \scalea{0.355} & \scalea{\subbst{0.169}} & \scalea{\subbst{0.214}} \\
    & \scalea{192} & \scalea{\bst{0.201}} & \scalea{\bst{0.241}} & \scalea{0.221} & \scalea{\subbst{0.254}} & \scalea{{0.223}} & \scalea{0.275}  & \scalea{0.218} & \scalea{0.287}& \scalea{0.227} & \scalea{{0.266}} & \scalea{0.227} & \scalea{0.297} & \scalea{0.242} & \scalea{0.298} & \scalea{0.236} & \scalea{0.294} & \scalea{0.275} & \scalea{0.345}  & \scalea{0.225} & \scalea{0.259}& \scalea{0.313} & \scalea{0.371} & \scalea{\subbst{0.216}} & \scalea{0.255} \\
    & \scalea{336} & \scalea{\bst{0.259}} & \scalea{\bst{0.283}} & \scalea{0.278} & \scalea{0.296} & \scalea{{0.272}} & \scalea{0.316}  & \scalea{\subbst{0.261}} & \scalea{0.321}& \scalea{0.283} & \scalea{{0.305}} & \scalea{0.278} & \scalea{0.346} & \scalea{0.287} & \scalea{0.335} & \scalea{0.282} & \scalea{0.332} & \scalea{0.338} & \scalea{0.379}  & \scalea{0.278} & \scalea{0.297}& \scalea{0.359} & \scalea{0.393} & \scalea{{0.271}} & \scalea{\subbst{0.293}}  \\
    & \scalea{720} & \scalea{\bst{0.337}} & \scalea{\bst{0.337}} & \scalea{0.358} & \scalea{0.347} & \scalea{\subbst{0.340}} & \scalea{0.363}  & \scalea{{0.348}} & \scalea{0.362}& \scalea{0.359} & \scalea{{0.355}} & \scalea{0.368} & \scalea{0.407} & \scalea{0.351} & \scalea{0.386} & \scalea{0.347} & \scalea{0.384} & \scalea{0.408} & \scalea{0.418}  & \scalea{0.354} & \scalea{0.348}& \scalea{0.440} & \scalea{0.446} & \scalea{0.350} & \scalea{\subbst{0.344}} \\
    \cmidrule(lr){2-26}
    & \scalea{Avg} & \scalea{\bst{0.237}} & \scalea{\bst{0.264}} & \scalea{0.258} & \scalea{0.278} & \scalea{{0.255}} & \scalea{0.299}  & \scalea{0.251} & \scalea{0.305}& \scalea{0.262} & \scalea{{0.288}} & \scalea{0.262} & \scalea{0.324} & \scalea{0.271} & \scalea{0.320} & \scalea{0.265} & \scalea{0.317} & \scalea{0.311} & \scalea{0.361}  & \scalea{0.259} & \scalea{0.281} & \scalea{0.349} & \scalea{0.391} & \scalea{\subbst{0.251}} & \scalea{\subbst{0.276}}\\

    \bottomrule
  \end{tabular}
\end{table*}

\subsection{Experiment Settings}
\label{sec:experimental setting}
In this section, we evaluate the efficacy of FSMLP on time series  forecasting. 
We show that our FSMLP can serve as a foundation model with competitive performance on these tasks. 

\paragraph{Datasets} Our study delves into the analysis of seven widely-used real-world multi-channel time series forecasting datasets. These datasets encompass diverse domains, including ECL Transformer Temperature (ETTh1, ETTh2, ETTm1, and ETTm2) \cite{informer}, ECL, Traffic, and Weather, as utilized by Autoformer \cite{wu2021autoformer}. For fairness in comparison, we adhere to a standardized protocol \cite{itransformer}, dividing all forecasting datasets into training, validation, and test sets. Specifically, we employ a ratio of 6:2:2 for the ETT dataset and 7:1:2 for the remaining datasets. Refer to Table \ref{tab:datasets} for an overview of the characteristics of these datasets.
\paragraph{Baselines} We compare FSMLP against a variety of state-of-the-art baselines.
Channel-independent methods include PatchTST, RLinear \cite{revin} and DLinear \cite{dlinear }  SCINet \cite{scinet}, FITS\cite{fits}. 
Channel-mix methods include    Crossformer \cite{crossformer}, FEDformer \cite{fedformer}, Autoformer \cite{wu2021autoformer}, iTransformer \cite{itransformer}, TSMixer \cite{chen2023tsmixerallmlparchitecturetime}, FreTS \cite{frets}, and FiLM \cite{film}.
\paragraph{Inplemental Details}
Similar to the settings in \cite{timesnet}, we set the look-back window to 96 for all datasets. Additionally, we incorporate the instance normalization block and reverse instance normalization \cite{revin}. All reported results are the averages over 10 random seeds. The baseline models used in this study were carefully reproduced with hyperparameters obtained from the TimesNet repository \cite{timesnet}, following reproducibility verification.
Training was conducted over 100 epochs, with early stopping applied and a patience of 10 epochs, as done in \cite{fits}. For all baseline models, the Adam optimizer \cite{adam} was used. To simplify the implementation, we use the Discrete Cosine Transform (DCT) as our frequency-domain transformation, as it operates solely on real numbers. Finally, we set the number of layers in our method to $3$, with a hidden dimension of $128$.

\subsection{Main Results}

The experimental results show that FSMLP outperforms several recent state-of-the-art models across various datasets in long-term forecasting tasks with forecast lengths of \(\tau = 96, 192, 336, 720\). While models such as FITS, iTransformer, FreTS, PatchTST, and others have shown promise, FSMLP demonstrates significant improvements, especially in datasets with varying complexities in channel dependencies.

For datasets with simpler channel dependencies, such as ETTm1 and ETTm2, FSMLP achieves notable improvements over both simpler models like FITS and more complex ones like iTransformer and PatchTST. For instance, FSMLP achieves an average MSE of \(0.365\) and MAE of \(0.382\) on ETTm1, outperforming all existing methods, including iTransformer.
iTransformer tends to struggle with overfitting, especially as the forecasting length increases. This issue can be attributed to the large number of parameters in its attention mechanism, which allows the model to overly fit to noise present in long-term dependencies.

In contrast, FSMLP effectively addresses this issue by incorporating Simplex-MLP, which constrains the weight space, reducing overfitting and enhancing generalization. This capability becomes particularly evident in datasets with more complex channel dependencies, such as ECL and Traffic, where FSMLP significantly outperforms other models. For instance, on the Traffic dataset, \textbf{FSMLP achieves the best average MSE of \(0.415\) and MAE of \(0.272\), surpassing models like FreTS, FITS, PatchTST, and iTransformer}.

A key limitation of FreTS is its exclusive reliance on FFT to capture both time and channel dependencies. While FFT can capture frequency-domain patterns effectively, it fails to model the complex inter-channel dependencies that are critical in datasets like Traffic and ECL, where channel interactions play a pivotal role. Similarly, FITS utilizes FFT for frequency-domain transformations and employs a single linear layer for time dependencies, but it lacks explicit modeling of channel dependencies, which diminishes its effectiveness on datasets with intricate channel interactions.

PatchTST, though leveraging attention mechanisms to capture time dependencies, falls short in modeling inter-channel dependencies. This limitation makes PatchTST less suitable for datasets like ECL and Traffic, where both time and channel dependencies are essential for accurate forecasting. FSMLP, by contrast, excels in capturing both time and channel dependencies within a unified framework, offering a distinct advantage over PatchTST and the other models.

On the other hand, FSMLP outperforms other state-of-the-art models on both simple and complex datasets by effectively modeling both time and channel dependencies. Its ability to regularize through Simplex-MLP and handle long-term forecasting tasks without overfitting makes it an ideal solution, particularly for complex datasets like Traffic and ECL, where intricate channel dependencies play a crucial role.

\subsection{Ablation Study}
We conduct an ablation study to analyze the individual contributions of different components in our proposed method. The results are presented in Table \ref{tab:ablation}, where we systematically remove key components while retaining others to evaluate their impact on the overall performance.

\textbf{First}, we examine the performance of our model without the Simplex-MLP constraint. The results show a significant performance drop across all datasets. This highlights the critical role of the Simplex-MLP component in regularizing the model, promoting a more structured representation, and ultimately improving generalization.

\textbf{Next}, we remove the Frequency Transformation(FT), a key component that allows our model to operate in the frequency domain. Without FT, our model performs suboptimally on datasets such as ETTh1 and Traffic, with MSE values significantly higher compared to the full model. This demonstrates the importance of capturing periodic patterns in the frequency domain to effectively model channel dependencies and reduce overfitting.

\textbf{Finally}, we evaluate the model without the frequency loss term. Similar to the Frequency Transformation ablation, removing frequency loss leads to a degradation in performance, especially for datasets such as Weather and ECL, where the model's ability to generalize is compromised. This confirms that incorporating frequency loss contributes to the model's ability to focus on relevant features and further mitigates overfitting.
\begin{table*}
    \small
    \centering
    \caption{The ablation experimental results. All results are averaged over forecasting lengths. w/o means that removing this component but retaining other components.}
     \label{tab:inference}
    \resizebox{\linewidth}{!}{\begin{tabular}{c|cl|cc|cl|ll|ll|ll|ll}
        \toprule
         &  \multicolumn{2}{c}{ETTh1}&  \multicolumn{2}{c}{ETTh2}&  \multicolumn{2}{c}{ETTm1}&\multicolumn{2}{c}{ETTm2
}& \multicolumn{2}{c}{Traffic}& \multicolumn{2}{c}{Weather
}& \multicolumn{2}{c}{ECL
}\\
 & MSE& MAE& MSE& MAE& MSE& MAE& MSE& MAE& MSE& MAE& MSE& MAE& MSE&MAE\\
         \midrule
         w/o Simplex-MLP&  0.478 &0.465&  0.397&  0.408&  0.408&  0.405&0.295 & 
0.336& 0.489& 0.310& 0.263& 0.281& 0.205
&0.289\\
         w/o Frequency Transformation&  0.422 &0.432&  0.359&  0.391&  0.379&  0.392&0.272 & 
0.320& 0.421& 0.281& 0.245& 0.272& 0.165
&0.261\\
         w/o Frequncy Loss&  0.420 &0.431&  0.355&  0.386&  0.368&  0.388&0.269 & 
0.316& 0.416& 0.276& 0.241
& 0.269& 0.163
&0.258\\
         \midrule
         Ours&  \bst{0.416} &\bst{0.425}& \bst{0.350}&  \bst{0.384}& \bst{ 0.365}&  \bst{0.382}&\bst{0.265} & \bst{0.311}& \bst{0.415}& \bst{0.272}& \bst{0.237}& \bst{0.264}& \bst{0.159}&\bst{0.252}\\
         \bottomrule
    \end{tabular}}
    
    \label{tab:ablation}
\end{table*}

\subsection{Efficiency Analysis}

In this section, we analyze the efficiency of FSMLP and baselines, the setting is as the same as the main results.

\paragraph{Inference.} 
We evaluate the inference time of our proposed FSMLP method and compare it with several state-of-the-art models, as shown in Table \ref{tab:inference}. The results, measured per 256 samples across various datasets, demonstrate the efficiency of FSMLP.

Our \textbf{FSMLP method achieves the fastest inference times on most datasets}, consistently outperforming the compared models. These results highlight FSMLP's efficiency, making it a suitable choice for real-time applications where low latency is crucial.
In contrast, models like Autoformer and TimesNet exhibit significantly higher inference times, particularly on larger datasets. The superior efficiency of FSMLP is attributed to its optimized architecture that effectively captures inter-channel dependencies with lower computational overhead, thereby reducing the overall inference time. This efficiency makes FSMLP not only accurate but also practical for deployment in environments where computational resources and time are limited.
\paragraph{Training.}

\begin{figure*}
    \centering
    \subfigure[ETTh1]{\includegraphics[width=0.48\linewidth]{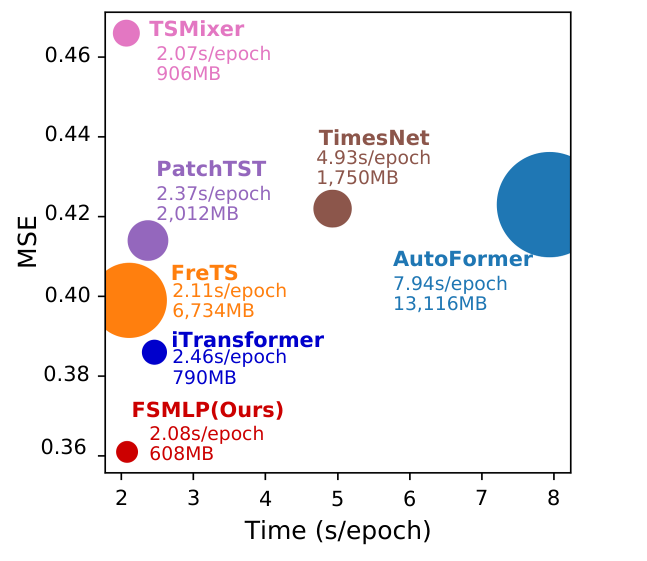}
   }
    \subfigure[Weather]{\includegraphics[width=0.48\linewidth]{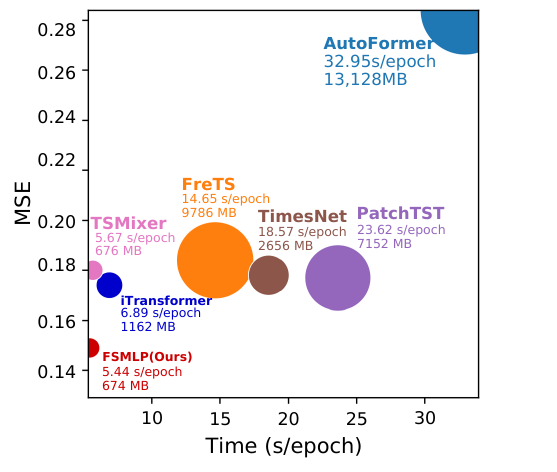}}
     \caption{This is the efficiency comparison of the proposed FSMLP with baselines on ETTh1 dataset with $L=96$ and $\tau=96$.}
     \label{fig:efficiency}

\end{figure*}

As shown in Fig. \ref{fig:efficiency}, FSMLP demonstrates both effectiveness and efficiency. \textbf{The framework not only achieves high forecasting accuracy but also offers significant computational advantages}. FSMLP requires considerably less memory and delivers faster training times compared to several state-of-the-art models, such as iTransformer, PatchTST, FreTS, TSMixer, AutoFormer, and TimesNet. These advantages make FSMLP a highly practical solution for time series forecasting, especially in resource-constrained environments where both memory usage and training speed are critical factors.

\begin{table*}
    \centering
        \caption{Inference time per $256$ samples. }
          \renewcommand{\arraystretch}{1.2} 
 \label{tab:inference}

{    \begin{tabular}{c|ccccccc}
    \toprule
         &  ETTh1&  ETTh2&  ETTm1&  ETTm2&  Weather&  ECL& Traffic\\
                  \midrule

         Autoformer&  0.219s&  0.217s&  0.220s&  0.215s&  0.659s&  1.271s& 2.836s\\
         TimesNet&  0.049s&  0.042s&  0.045s&  0.047s&  0.244s&  2.504s& 2.536s\\
         iTransformer&  0.021s&  0.020s&  0.021s&  0.024s&  0.061s&  0.121s& 0.203s\\
         PatchTST&  0.018s&  0.019s&  0.025s&  0.028s&  0.134s&  0.207s& 0.368s\\
         FreTS&  0.019s&  0.017s&  0.022s&  0.022s&  0.023&  0.069s& 0.105s\\
         TSMixer&  0.019s&  0.018s&  0.023s&  0.021&  0.021s&  0.081s& 0.127s\\
         \midrule
         FSMLP(Ours)&  \textbf{0.018s}&  \textbf{0.017}s&  \textbf{0.022s}&  \textbf{0.020s}&  \textbf{0.021s}&  \textbf{0.064s}& \textbf{0.106s}\\
         \bottomrule
    \end{tabular}}
   
\end{table*}

\subsection{Analysis of Simplex-MLP}
\subsubsection{Different Implementations of Simplex-MLP}
In this section we compare three implementations of Simplex-MLP.
As the results shown in Table \ref{tab:implementations}, it is evident that the \textbf{Logarithm implementation of Simplex-MLP outperforms both the Absolute Value and Square implementations in terms of MSE and MAE across all datasets.} 
Specifically, the Logarithm implementation achieves an MSE of 0.416 and an MAE of 0.425 on the ETTh1 dataset, which is superior to the other two implementations. Similar improvements are observed in the ETTh2, ETTm1, ETTm2, and Weather datasets. The superior performance of the Logarithm implementation can be attributed to its ability to better capture the underlying dependencies in the data while maintaining numerical stability. This results in a more accurate and reliable forecasting model, as evidenced by the consistently lower error metrics. In conclusion, the Logarithm implementation of Simplex-MLP demonstrates its effectiveness in handling different time series datasets, making it a preferable choice for capturing complex dependencies and enhancing overall model performance.

\subsubsection{Simplex MLP for other methods}
\label{sec:Analysis of Simplex-MLP Improvements}
Table \ref{tab:improve} illustrates the improvements achieved by \textbf{incorporating Simplex-MLP into TSMixer and Autoformer} across various datasets. The results highlight the significant role of Simplex-MLP in enhancing model performance, particularly by reducing overfitting and improving generalization.

For instance, for TSMixer, the integration of Simplex-MLP yields noticeable performance gains across all datasets. The most significant improvement is observed on the ETTh2 dataset, where the model's ability to capture complex inter-channel dependencies is greatly enhanced. This suggests that Simplex-MLP, by constraining the weight space to the Standard \(n\)-simplex, promotes a more structured and compact representation of the data, which helps mitigate the overfitting observed in the original TSMixer model. On the other datasets, such as ETTh1, ETTm1, and Traffic, Simplex-MLP also contributes to reducing prediction errors, with consistent gains in both MSE and MAE, highlighting its effectiveness across diverse time series forecasting tasks.


\begin{table*}
    \centering
     \caption{Improvement of Autoformer and TSMixer with Simplex-MLP.}
         \label{tab:improve}

    \begin{tabular}{c|ccccccclll}
    \toprule
         &  \multicolumn{2}{c}{ETTh1}&  \multicolumn{2}{c}{ETTh2}&  \multicolumn{2}{c}{ETTm1}& \multicolumn{2}{c}{ETTm2}& \multicolumn{2}{c}{Traffic}\\
 & MSE& MAE& MSE& MAE& MSE& MAE& MSE& MAE& MSE& MAE\\
 \midrule
         TSMixer&  0.623&  0.585&  2.025&  1.194&  0.529&  0.513&  1.030& 0.753& 0.573& 0.388\\
         TSMixer(Simplex Version)&  \textbf{0.553}&  \textbf{0.530}&  \textbf{0.589}&  \textbf{0.534}&  \textbf{0.442}&  \textbf{0.459}&  \textbf{0.366}& \textbf{0.406}& \textbf{0.525}& \textbf{0.347}\\
         \midrule
 \textit{Improvement}&
\textcolor{abl}{7.00\%$\uparrow$} & \textcolor{abl}{5.50\%$\uparrow$} & \textcolor{abl}{143.60\%$\uparrow$} & \textcolor{abl}{66.00\%$\uparrow$} & \textcolor{abl}{8.70\%$\uparrow$} & \textcolor{abl}{5.40\%$\uparrow$} & \textcolor{abl}{66.40\%$\uparrow$} & \textcolor{abl}{34.70\%$\uparrow$} & \textcolor{abl}{4.80\%$\uparrow$} & \textcolor{abl}{4.10\%$\uparrow$} \\

\midrule
         Autoformer&  0.476&  0.477&  0.478&  0.483&  0.596&  0.517&  0.326& 0.366& 1.001& 0.652\\
         Autoformer(Simplex Version)&  \textbf{0.434}&  \textbf{0.462}&  \textbf{0.439}&  \textbf{0.453}&  \textbf{0.570}&  \textbf{0.511}&  \textbf{0.319}& \textbf{0.361}& \textbf{0.631}& \textbf{0.388}\\
                  \midrule

 \textit{Improvement}& \textcolor{abl}{4.20\%$\uparrow$}&\textcolor{abl}{1.50\%$\uparrow$}& \textcolor{abl}{3.90\%$\uparrow$}& \textcolor{abl}{3.00\%$\uparrow$}& \textcolor{abl}{2.60\%$\uparrow$}& \textcolor{abl}{0.60\%$\uparrow$}& \textcolor{abl}{0.70\%$\uparrow$}& \textcolor{abl}{0.50\%$\uparrow$}& \textcolor{abl}{37.00\%$\uparrow$}& \textcolor{abl}{26.40\%$\uparrow$}
\\
\bottomrule
    \end{tabular}
   
\end{table*}
\begin{table*}
    \centering
        \caption{Analysis of different kinds of implementations of Simplex-MLP.}
  \renewcommand{\arraystretch}{1.2} 

    {\begin{tabular}{c|cc|cc|cc|cc|cl}
    \toprule
         &  \multicolumn{2}{c}{ETTh1}&  \multicolumn{2}{c}{ETTh2}&  \multicolumn{2}{c}{ETTm1}&  \multicolumn{2}{c}{ETTm2}&  \multicolumn{2}{c}{Weather}\\
 & MSE& MAE& MSE& MAE& MSE& MAE& MSE& MAE& MSE&MAE\\
 \midrule
         Absolute Value &  0.421&  0.433&  0.356&  0.386&  0.365&  0.384&  0.269&  0.315&  0.241&0.266\\
         Square&  0.419&  0.428&  0.353&  0.387&  0.367&  0.385&  0.267&  0.313&  0.239&0.267\\
         \midrule
         Log&  \textbf{0.416}&  \textbf{0.425}&  \textbf{0.350}&  \textbf{0.384}&  \textbf{0.365}&  \textbf{0.382}&  \textbf{0.265}&\textbf{ 0.311}&  \textbf{0.237}&\textbf{0.264}\\
         \bottomrule
    \end{tabular}}
    \label{tab:implementations}
\end{table*}
\subsubsection{Comparison With Other Constraints}
In this section, we compare our Simplex-MLP with other constraints aimed at reducing overfitting, including L1 Norm, L2 Norm, and Compressed MLP.
\textit{L1 Norm regularization}, also known as Lasso regularization, adds the absolute value of the weights to the loss function:
\[\text{Loss} = \text{Loss}_{\text{original}} + \lambda \sum_{i} |w_i| \]
\textit{L2 Norm regularization}, also known as Ridge regularization, adds the squared value of the weights to the loss function:
\[ \text{Loss} = \text{Loss}_{\text{original}} + \lambda \sum_{i} w_i^2 \]
\textit{Compressed MLP} is a type of MLP that uses singular value decomposition (SVD) \[W = U \Sigma V^T\]. Specifically, compressed MLP only stores the largest \(k\) singular values and their corresponding vectors. During training, we first reconstruct the weight matrix from these \(k\) singular values and vectors before performing the forward pass.
In this experiment, we replace the Simplex-MLP in FSMLP with each of the aforementioned constraints and compare their performances with the original Simplex-MLP version.

From Table \ref{tab:overfitting}, we observe that Simplex-MLP consistently outperforms L1 Norm, L2 Norm, and Compressed MLP in terms of both MSE and MAE across all datasets. Specifically, on the ETTh1 dataset, Simplex-MLP achieves an MSE of 0.416 and MAE of 0.425, which is \textbf{significantly better than Compressed MLP (MSE = 0.461, MAE = 0.459), L2 Norm (MSE = 0.472, MAE = 0.463), and L1 Norm (MSE = 0.465, MAE = 0.463)}. Similarly, for the ECL dataset, Simplex-MLP achieves an MSE of 0.159 and MAE of 0.252, outperforming the other methods by a notable margin.

While Compressed MLP shows competitive results, especially on datasets like ETTh2 and ETTm2, it still falls behind Simplex-MLP in all cases, particularly with respect to the MAE metric. This suggests that the use of singular value decomposition (SVD) in Compressed MLP does not lead to superior performance when compared to the regularization-based methods or our proposed Simplex-MLP.

The Simplex-MLP demonstrates superior performance over traditional regularization methods and compressed MLP techniques in reducing overfitting and improving accuracy. Its ability to constrain the weight space within a well-defined standard simplex enables it to capture inter-channel dependencies effectively while mitigating the impact of extreme values, leading to consistent improvements across various time series forecasting tasks. These results underscore the effectiveness of Simplex-MLP in enhancing the performance of MLP-based models for time series forecasting, making it a valuable addition to the toolkit for handling complex datasets.

\begin{table*}
    \centering
        \caption{Comparison of Simplex MLP with other constraints. }
  \renewcommand{\arraystretch}{1.2} 

    \begin{tabular}{c|cc|cc|cc|cc|cc|cc|cc}
    \toprule
         &  \multicolumn{2}{c}{ETTh1}&  \multicolumn{2}{c}{ETTh2}&  \multicolumn{2}{c}{ETTm1}&  \multicolumn{2}{c}{ETTm2}&  \multicolumn{2}{c}{ECL} &  \multicolumn{2}{c}{Traffic}&\multicolumn{2}{c}{Weather}\\
Techniques& MSE& MAE& MSE& MAE& MSE& MAE& MSE& MAE& MSE&MAE &  MSE& MAE&MSE&MAE\\
          \midrule
         Compressed MLP&  0.461&  0.459&  0.368&  0.389&  0.379&  0.393&  0.275&  0.319&  0.176&0.271 &  0.454& 0.306&0.264&0.282\\
         L2 Norm&  0.472&  0.463&  0.389&  0.396&  0.392&  0.401&  0.289&  0.329&  0.184&0.275 &  0.468& 0.319&0.273&0.288\\
         L1 Norm&  0.465&  0.463&  0.382&  0.393&  0.385&  0.397&  0.286&  0.327&  0.181&0.271 &  0.455& 0.308&0.271&0.285\\
        \midrule
         Simplex MLP&  \textbf{0.416}&  \textbf{0.425}&  \textbf{0.350}&  \textbf{0.384}&  \textbf{0.365}&  \textbf{0.382}&  \textbf{0.265}&  \textbf{0.311}&  \textbf{0.159}&\textbf{0.252} &  \textbf{0.415}& \textbf{0.272}&\textbf{0.237}&\textbf{0.264}\\
         \bottomrule
    \end{tabular}
    \label{tab:overfitting}
\end{table*}

\subsection{Scalability}
\subsubsection{Partial Sample Training}

Table \ref{tab:partital-sample} presents the results of the scalability analysis using partial sample training, where both input and prediction lengths are set to 96. The performance of four models—FSMLP, FreTS, TSMixer, and FITS—was evaluated across various datasets, including ETTh1, ETTh2, ETTm1, ETTm2, and Weather. The results indicate that, generally, the models maintain stable performance as the proportion of training data increases.
FSMLP demonstrates consistent improvements in performance as more training data is provided, showing enhanced accuracy with additional data. This highlights its scalability and ability to leverage larger training datasets effectively. 
Moreover, FSMLP consistently shows better performances than the other methods under different training data scale.
These findings suggest that FSMLP is highly scalable and well-suited for real-world applications where the volume of training data can vary significantly.

\begin{table}
    \centering
        \caption{Scalability analysis using partial sample training results. Both the input length and the prediction length are $96$.}
\centering
  \renewcommand{\arraystretch}{1.2} 

\resizebox{\linewidth}{!}{
    \begin{tabular}{c|c|cc|cc|cc|cc}
    \toprule

        \multicolumn{2}{c}{}&  \multicolumn{2}{c}{FSMLP}&  \multicolumn{2}{c}{FreTS}&  \multicolumn{2}{c}{TSMixer}&  \multicolumn{2}{c}{FITS}\\
          
          &&  MSE&  MAE&  MSE&  MAE&  MSE&  MAE&  MSE&  MAE\\
          \midrule
          &\textbf{\textit{20\%}}&  \bst{0.412}&  \bst{0.419}&  0.690&  0.561&  0.890&  0.664&  0.459&  0.446\\
          &\textbf{\textit{40\%}}&  \bst{0.403}&  \bst{0.407}&  0.472&  0.464&  0.808&  0.676&  0.449&  0.439\\
          ETTh1&\textbf{\textit{60\%}}&  \bst{0.389}& \bst{0.399}&  0.442&  0.449&  0.598&  0.567&  0.438&  0.436\\
 & \textbf{\textit{80\%}}& \bst{0.373}& \bst{0.388}& 0.425& 0.436& 0.513& 0.514& 0.391&0.407\\
 & \textbf{\textit{100\%}}& \bst{0.361}& \bst{0.384}& 0.399& 0.412& 0.466& 0.482& 0.387&0.394\\
          \midrule
          &\textbf{\textit{20\%}}
&  \bst{0.305}&\bst{0.353}&  1.11&  0.73&  2.420&  1.17&  0.306&  0.359\\
          &\textbf{\textit{40\%}}
&  \bst{0.290}&  \bst{0.341}&  0.491&  0.477&  1.067&  0.77&  0.304&  0.348\\
          ETTh2&\textbf{\textit{60\%}}
&  \bst{0.283}& \bst{0.336}&  0.474&  0.469&  1.359&  0.887&  0.302&  0.354\\
 & \textbf{\textit{80\%}}
& \bst{0.282}& \bst{0.336}& 0.498& 0.475& 1.198& 0.824& 0.299&0.352\\
 & \textbf{\textit{100\%}}& \bst{0.277}& \bst{0.328}& 0.350& 0.403& 1.056& 0.806& 0.296&0.340\\
          \midrule
          &\textbf{\textit{20\%}}
&\bst{0.481}& \bst{0.448}&  1.028&  0.684&  0.792&  0.644&  0.677&  0.505\\
          &\textbf{\textit{40\%}}
& \bst{0.495}& \bst{0.442}&  0.689&  0.546&  0.794&  0.632&  0.563&  0.479\\
 ETTm1& \textbf{\textit{60\%}}
&\bst{0.362}& \bst{0.386}& 0.487& 0.461& 0.675& 0.562& 0.532& 0.483\\
 & \textbf{\textit{80\%}}
& \bst{0.309}& \bst{0.346}& 0.393& 0.416& 0.544& 0.509& 0.428&0.420\\
 & \textbf{\textit{100\%}}& \bst{0.303}& \bst{0.342}& 0.339& 0.374& 0.479& 0.470& 0.365&0.380\\
 \midrule
 & \textbf{\textit{20\%}}
& \bst{0.183}& \bst{0.261}& 0.628& 0.554& 1.451& 0.877& 0.281& 0.344\\
 & \textbf{\textit{40\%}}
&\bst{0.177}&\bst{0.255}& 0.640& 0.564& 0.602& 0.568& 0.257& 0.326\\
 ETTm2& \textbf{\textit{60\%}}
& \bst{0.173}&\bst{0.252}& 0.428& 0.455& 0.464& 0.503& 0.217& 0.315\\
 & \textbf{\textit{80\%}}
& \bst{0.171}& \bst{0.249}& 0.303& 0.390& 0.339& 0.423& 0.201&0.287\\
 & \textbf{\textit{100\%}}& \bst{0.166}& \bst{0.247}& 0.190& 0.282& 0.250& 0.366& 0.186&0.269\\
 \midrule

 & \textbf{\textit{20\%}}
&\bst{0.161}& \bst{0.204}& 0.268& 0.296& 0.384& 0.407& 0.189& 0.233\\
 & \textbf{\textit{40\%}}
&\bst{0.152}&\bst{0.194}& 0.254& 0.284& 0.314& 0.334& 0.192& 0.231\\
 Weather& \textbf{\textit{60\%}}
& \bst{0.154}&\bst{0.196}& 0.221& 0.276& 0.284& 0.314& 0.191& 0.232\\
 & \textbf{\textit{80\%}}
& \bst{0.154}&\bst{0.199}& 0.205& 0.257& 0.244& 0.286& 0.191&0.231\\
 & \textbf{\textit{100\%}}& \bst{0.149}& \bst{0.193}& 0.184& 0.239& 0.180& 0.252& 0.169&0.214\\
   \bottomrule

    \end{tabular}}
    \label{tab:partital-sample}
\end{table}
\begin{table}
    \centering
        \caption{Comparison under longer input length. The results are averaged over prediction lengths.}
    \label{tab:input_len}
  \renewcommand{\arraystretch}{1.2} 

    \resizebox{\linewidth}{!}{\begin{tabular}{cc|cc|cc|cc|cc}
    \toprule
        \centering

          &&  \multicolumn{2}{c}{FSMLP}&  \multicolumn{2}{c}{FreTS}&  \multicolumn{2}{c}{FITS}&  \multicolumn{2}{c}{PatchTST}\\
          
          &&  MSE&  MAE&  MSE&  MAE&  MSE&  MAE&  MSE&  MAE\\
 \midrule
 & 192& \bst{0.155}&\bst{ 0.247}& 0.182& 0.284& 0.182& 0.283& 0.185& 0.279\\
 ECL& 336&\bst {0.153}&\bst{ 0.246}& 0.175& 0.279& 0.168& 0.272& 0.162& 0.263\\
 & 720& \bst{0149}&\bst{0.240}& 0.178& 0.281& 0.163& 0.265& 0.159& 0.261\\
 \midrule

 & 192&\bst{0.397}& \bst{0.266}& 0.515& 0.326& 0.473& 0.407& 0.431& 0.293\\
 Traffic& 336&\bst{0.389}&\bst{0.261}& 0.483& 0.317& 0.431& 0.295& 0.401& 0.278\\
 & 720&\bst {0.385}&\bst{0.258}& 0.458& 0.309& 0.411& 0.285& 0.395& 0.274\\
  \bottomrule
    \end{tabular}}
\end{table}

\begin{table*}
    \centering
        \caption{Analysis of different learning rates. All results are averaged over forecasting lengths.}
    \label{tab:learning_rate}
  \renewcommand{\arraystretch}{1.2} 

    \begin{tabular}{c|cc|cc|cc|cc|cl}
    \toprule
         &  \multicolumn{2}{c}{ETTh1}&  \multicolumn{2}{c}{ETTh2}&  \multicolumn{2}{c}{ETTm1}&  \multicolumn{2}{c}{ETTm2}&  \multicolumn{2}{c}{ECL}\\
Learning Rate & MSE& MAE& MSE& MAE& MSE& MAE& MSE& MAE& MSE&MAE\\
          \midrule

         0.02&  0.424&  0.431&  0.361&  0.393&  0.375&  0.387&  0.271&  0.317&  0.167&0.258\\
         0.01&  \textbf{0.416}&  \textbf{0.425}&  \textbf{0.350}&  \textbf{0.384}&  \textbf{0.365}&  \textbf{0.382}&  \textbf{0.265}&  \textbf{0.311}&  \textbf{0.159}&\textbf{0.252}\\
         0.005&  0.417&  0.425&  0.351&  0.384&  0.365&  0.383&  0.268&  0.313&  0.159&0.252\\
         0.001&  0.421&  0.428&  0.356&  0.388&  0.367&  0.385&  0.269&  0.314&  0.166&0.259\\
         \bottomrule
    \end{tabular}
\end{table*}

\begin{table*}
    \centering
        \caption{Analysis of different Batch Sizes All results are averaged over forecasting lengths.}
    \label{tab:batch_size}
  \renewcommand{\arraystretch}{1.2} 

    \begin{tabular}{c|cc|cc|cc|cc|cl}
    \toprule
         &  \multicolumn{2}{c}{ETTh1}&  \multicolumn{2}{c}{ETTh2}&  \multicolumn{2}{c}{ETTm1}&  \multicolumn{2}{c}{ETTm2}&  \multicolumn{2}{c}{ECL}\\
Batch Size& MSE& MAE& MSE& MAE& MSE& MAE& MSE& MAE& MSE&MAE\\
          \midrule

         512&  0.424&  0.431&  0.357&  0.389&  0.372&  0.389&  0.273&  0.319&  0.167&0.258\\
         256&  \textbf{0.416}&  \textbf{0.425}&  \textbf{0.350}&  \textbf{0.384}&  \textbf{0.365}&  \textbf{0.382}&  \textbf{0.265}&  \textbf{0.311}&  \textbf{0.159}&\textbf{0.252}\\
         128&  0.419&  0.427&  0.353&  0.386&  0.367&  0.384&  0.267&  0.313&  0.159&0.252\\
         64&  0.420&  0.427&  0.352&  0.386&  0.368&  0.384&  0.268&  0.314&  0.162&0.254\\
         \bottomrule
    \end{tabular}
\end{table*}
\subsubsection{Large Dataset}
The Traffic dataset, known for its large scale and complexity, presents a significant challenge for time series forecasting models due to its high dimensionality and the intricate dependencies between different traffic sensors. Our FSMLP method achieves outstanding performance on this dataset, demonstrating its ability to effectively manage and process large-scale datasets. The results show that FSMLP consistently outperforms other models in terms of MSE and MAE, highlighting its robustness and scalability. This performance indicates that FSMLP can generalize well and maintain high accuracy even with the increased data volume and complexity inherent in large-scale datasets, making it a suitable choice for real-world applications that require handling extensive time series data.

\subsubsection{Longer Prediction Lengths}
\begin{table}
    \centering
        \caption{Longer prediction length comparison.}    
        \label{tab:varing_prediction_len}
  \renewcommand{\arraystretch}{1.2} 
\resizebox{\linewidth}{!}{
    \begin{tabular}{cc|cc|cc|cc|cc}
    \toprule
        \centering

          &&  \multicolumn{2}{c}{FSMLP}&  \multicolumn{2}{c}{FreTS}&  \multicolumn{2}{c}{TSMixer}&  \multicolumn{2}{c}{PatchTST}\\
          
          &&  MSE&  MAE&  MSE&  MAE&  MSE&  MAE&  MSE&  MAE\\
          \midrule
          &960&  \textbf{0.521}&  \textbf{0.495}&  0.653&  0.587&  0.814&  0.707&  0.542&  0.507\\
          ETTh1&1440&  \textbf{0.626}&  \textbf{0.557}&  0.778&  0.658&  0.956&  0.782&  0.640&  0.569\\
          &2160&  \textbf{0.813}& \textbf{0.648}&  0.969&  0.752&  1.114&  0.855&  0.842&  0.662\\
          \midrule
          &960&  \textbf{0.435} &\textbf{0.456}&  1.060&  0.745&  2.690&  1.413&  0.484&  0.484\\
          ETTh2&1440&  \textbf{0.532}&  \textbf{0.516}&  1.467&  0.880&  2.884&  1.484&  0.551&  0.525\\
          &2160&  \textbf{0.553}&  \textbf{0.527}&  1.555&  0.879&  3.018&  1.584&  0.586&  0.540\\
          \midrule
          &960&\textbf{0.478}&  \textbf{0.448}&  0.540&  0.504&  0.684&  0.605&  0.494&  0.458\\
          ETTm1&1440& \textbf{0.505}& \textbf{0.465}&  0.592&  0.537&  0.755&  0.650&  0.525&  0.479\\
 & 2160&\textbf{0.511}& \textbf{0.469}& 0.635& 0.569& 0.819& 0.691& 0.532& 0.483\\
 \midrule
 & 960& \textbf{0.433}& \textbf{0.415}& 0.628& 0.554& 1.825& 1.125& 0.451& 0.429\\
 ETTm2& 1440& \textbf{0.459}& \textbf{0.439}& 0.640& 0.564& 1.872& 1.116& 0.473& 0.453\\
 & 2160& \textbf{0.449}&\textbf{0.443}& 0.734& 0.622& 1.891& 1.136& 0.478& 0.461\\
 \midrule

 & 960&\textbf{0.368}& \textbf{0.401}& 0.379& 0.412& 0.384& 0.407& 0.388& 0.370\\
 Weather& 1440&\textbf{0.381}&\textbf{0.397}& 0.394& 0.420& 0.409& 0.436& 0.419& 0.389\\
 & 2160& \textbf{0.402}&\textbf{0.428}& 0.417& 0.434& 0.415& 0.448& 0.457& 0.413\\
  \bottomrule
    \end{tabular}}
\end{table}
Table \ref{tab:varing_prediction_len} illustrates the superior performance of FSMLP compared to FreTS, TSMixer, and PatchTST across multiple datasets and increasing prediction lengths. FSMLP's consistent performance, even with extended prediction lengths, highlights its scalability and resilience to overfitting. For datasets like ETTh1 and ETTh2, FSMLP consistently achieves the lowest MSE and MAE values across all tested prediction lengths (960, 1440, and 2160), indicating its capability to handle larger forecasting windows effectively. Similar trends are observed in ETTm1 and ETTm2 datasets, where FSMLP maintains lower errors consistently, showcasing its ability to capture long-term dependencies without overfitting. The Simplex-MLP constraints, which limit weights within the standard \(n\)-simplex, reduce the influence of redundant noise and enhance generalization. This advantage is further confirmed in the Weather dataset, where FSMLP achieves the lowest error rates, effectively handling complex temporal patterns without overfitting. The combination of frequency domain transformations and Simplex-MLP regularization significantly contributes to FSMLP's robust performance.

\subsubsection{Longer Input Lengths}

Table \ref{tab:input_len} provides a comparative analysis of FSMLP, FreTS, FITS, and PatchTST across different input lengths (192, 336, and 720) on the ECL and Traffic datasets. The results, averaged over the prediction lengths, demonstrate that FSMLP consistently achieves the best performance across all input lengths for both datasets. For the ECL dataset, FSMLP shows remarkable stability and superiority, maintaining the lowest error rates compared to other models, which exhibit higher error rates. This highlights FSMLP's ability to effectively utilize longer input sequences without overfitting. In the complex and large-scale Traffic dataset, FSMLP also demonstrates robust performance, consistently outperforming FreTS, FITS, and PatchTST across all input lengths. This underscores FSMLP's scalability and adaptability, effectively capturing long-term dependencies and making it a reliable choice for diverse time series forecasting tasks.

\subsection{Complexity}

\begin{table}
    \centering
    \caption{Comparison of the computational complexity of FSMLP and other baselines. \( P \) represents the patch size of PatchTST, \( L \) represents the input length, and \( N \) represents the number of channels.}
    \renewcommand{\arraystretch}{1.2}
    \resizebox{\linewidth}{!}{
    \begin{tabular}{cccccc}
    \toprule
         FSMLP & iTransformer & PatchTST & FITS & FreTS & TSMixer \\
       \(O(NL)\) & \(O(N^2L)\) & \(O\left(\frac{L^2}{P^2}N\right)\) & \(O(NL)\) & \(O(NL)\) & \(O(NL)\) \\
    \bottomrule
    \end{tabular}}
    \label{tab:complexity_comparison}
\end{table}

Table \ref{tab:complexity_comparison} compares the computational complexity of FSMLP with other state-of-the-art models. FSMLP, FITS, FreTS, and TSMixer all have a linear complexity of \( O(NL) \), making them scalable and efficient for large datasets. In contrast, iTransformer has a higher complexity of \( O(N^2L) \) due to its use of attention mechanisms to capture inter-channel dependencies, which can lead to higher computational costs and potential overfitting. PatchTST's complexity, \( O\left(\frac{L^2}{P^2}N\right) \), reflects its use of patches to model time dependencies. FSMLP's linear complexity, combined with its ability to effectively capture both time and channel dependencies, highlights its efficiency and suitability for large-scale time series forecasting tasks.

\subsection{Hyperparameter Sensitivity Analysis}
\paragraph{Learning Rate.} The analysis of different learning rates is summarized in Table \ref{tab:learning_rate}. As observed, the differences in model performance across the various learning rates are relatively small. The MSE and MAE values show only slight variation between learning rates of 0.02, 0.01, 0.005, and 0.001. This suggests that the model's performance remains fairly stable across the tested learning rates, with no dramatic changes in forecasting accuracy. 

\paragraph{Batch Size.} The analysis of different batch sizes is summarized in Table \ref{tab:batch_size}. As shown, the model's performance does not exhibit significant variation across different batch sizes.  This suggests that the model's performance is not highly sensitive to the choice of batch size within the tested range. Although there are small fluctuations, the overall results indicate that different batch sizes provide comparable forecasting accuracy. This indicates that the model can maintain stable performance across a range of batch sizes, offering flexibility in terms of computational efficiency without significantly impacting the predictive outcomes.

\vspace{3pt}
\section{Conclusion}
In this work, we introduce the FSMLP, a novel framework designed to address the challenges of modeling inter-channel dependencies and mitigating overfitting in time series forecasting. By constraining the model weights to the Standard \(n\)-Simplex, FSMLP enforces a geometric structure that regularizes the learning process, reducing the influence of redundant noise and leading to improved generalization. The use of frequency domain transformations further enhances the model’s ability to capture periodic dependencies across channels. Experimental results across multiple benchmark datasets demonstrate that FSMLP consistently outperforms state-of-the-art methods, especially in large-scale, long-term forecasting tasks, highlighting its scalability and robustness. Our findings suggest that FSMLP offers a promising solution for more efficient and reliable time series forecasting, with potential applications in diverse domains such as energy consumption, web data analysis, and weather prediction.

\vspace{3pt}
\section*{Acknowledgments}
This work is supported by Chinese National Natural Science Foundation Projects (Grant No.62206259) and the Fundamental Research Funds for the Central Universities.

\vspace{3pt}
\bibliography{references}

@misc{adam,
      title={Adam: A Method for Stochastic Optimization}, 
      author={Diederik P. Kingma and Jimmy Ba},
      year={2017},
      eprint={1412.6980},
      archivePrefix={arXiv},
      primaryClass={cs.LG},
      url={https://arxiv.org/abs/1412.6980}, 
}

@article{patchtst,
  title={A time series is worth 64 words: Long-term forecasting with transformers},
  author={Nie, Yuqi and Nguyen, Nam H and Sinthong, Phanwadee and Kalagnanam, Jayant},
  journal={arXiv preprint arXiv:2211.14730},
  year={2022}
}

@misc{xue2024card,
      title={CARD: Channel Aligned Robust Blend Transformer for Time Series Forecasting}, 
      author={Wang Xue and Tian Zhou and Qingsong Wen and Jinyang Gao and Bolin Ding and Rong Jin},
      year={2024},
      eprint={2305.12095},
      archivePrefix={arXiv},
      primaryClass={cs.LG}
}

@misc{liu2024timer,
      title={Timer: Transformers for Time Series Analysis at Scale}, 
      author={Yong Liu and Haoran Zhang and Chenyu Li and Xiangdong Huang and Jianmin Wang and Mingsheng Long},
      year={2024},
      eprint={2402.02368},
      archivePrefix={arXiv},
      primaryClass={cs.LG}
}

@misc{oreshkin2020nbeats,
      title={N-BEATS: Neural basis expansion analysis for interpretable time series forecasting}, 
      author={Boris N. Oreshkin and Dmitri Carpov and Nicolas Chapados and Yoshua Bengio},
      year={2020},
      eprint={1905.10437},
      archivePrefix={arXiv},
      primaryClass={cs.LG}
}

@misc{lstn,
      title={Modeling Long- and Short-Term Temporal Patterns with Deep Neural Networks}, 
      author={Guokun Lai and Wei-Cheng Chang and Yiming Yang and Hanxiao Liu},
      year={2018},
      eprint={1703.07015},
      archivePrefix={arXiv},
      primaryClass={cs.LG}
}

@misc{liu2023nonstationary,
      title={Non-stationary Transformers: Exploring the Stationarity in Time Series Forecasting}, 
      author={Yong Liu and Haixu Wu and Jianmin Wang and Mingsheng Long},
      year={2023},
      eprint={2205.14415},
      archivePrefix={arXiv},
      primaryClass={cs.LG}
}

@misc{foundationsurvey,
      title={A Survey of Time Series Foundation Models: Generalizing Time Series Representation with Large Language Model}, 
      author={Jiexia Ye and Weiqi Zhang and Ke Yi and Yongzi Yu and Ziyue Li and Jia Li and Fugee Tsung},
      year={2024},
      eprint={2405.02358},
      archivePrefix={arXiv},
      primaryClass={cs.LG}
}

@inproceedings{fre,
  title={Learning in the frequency domain},
  author={Xu, Kai and Qin, Minghai and Sun, Fei and Wang, Yuhao and Chen, Yen-Kuang and Ren, Fengbo},
  booktitle={Proceedings of the IEEE/CVF conference on computer vision and pattern recognition},
  pages={1740--1749},
  year={2020}
}

@misc{atfnet,
      title={ATFNet: Adaptive Time-Frequency Ensembled Network for Long-term Time Series Forecasting}, 
      author={Hengyu Ye and Jiadong Chen and Shijin Gong and Fuxin Jiang and Tieying Zhang and Jianjun Chen and Xiaofeng Gao},
      year={2024},
      eprint={2404.05192},
      archivePrefix={arXiv},
      primaryClass={cs.LG}
}

@misc{samformer,
      title={Unlocking the Potential of Transformers in Time Series Forecasting with Sharpness-Aware Minimization and Channel-Wise Attention}, 
      author={Romain Ilbert and Ambroise Odonnat and Vasilii Feofanov and Aladin Virmaux and Giuseppe Paolo and Themis Palpanas and Ievgen Redko},
      year={2024},
      eprint={2402.10198},
      archivePrefix={arXiv},
      primaryClass={cs.LG}
}

@inproceedings{informer,
  author    = {Haoyi Zhou and
               Shanghang Zhang and
               Jieqi Peng and
               Shuai Zhang and
               Jianxin Li and
               Hui Xiong and
               Wancai Zhang},
  title     = {Informer: Beyond Efficient Transformer for Long Sequence Time-Series Forecasting},
  booktitle = {The Thirty-Fifth {AAAI} Conference on Artificial Intelligence, {AAAI} 2021, Virtual Conference},
  volume    = {35},
  number    = {12},
  pages     = {11106--11115},
  publisher = {{AAAI} Press},
  year      = {2021},
}

@misc{ni2024basisformer,
      title={BasisFormer: Attention-based Time Series Forecasting with Learnable and Interpretable Basis}, 
      author={Zelin Ni and Hang Yu and Shizhan Liu and Jianguo Li and Weiyao Lin},
      year={2024},
      eprint={2310.20496},
      archivePrefix={arXiv},
      primaryClass={cs.LG}
}

@ARTICLE{difformer,
  author={Li, Bing and Cui, Wei and Zhang, Le and Zhu, Ce and Wang, Wei and Tsang, Ivor W. and Zhou, Joey Tianyi},
  journal={IEEE Transactions on Pattern Analysis and Machine Intelligence}, 
  title={DifFormer: Multi-Resolutional Differencing Transformer With Dynamic Ranging for Time Series Analysis}, 
  year={2023},
  volume={45},
  number={11},
  pages={13586-13598},
  doi={10.1109/TPAMI.2023.3293516}}

@misc{tft,
      title={Temporal Fusion Transformers for Interpretable Multi-horizon Time Series Forecasting}, 
      author={Bryan Lim and Sercan O. Arik and Nicolas Loeff and Tomas Pfister},
      year={2020},
      eprint={1912.09363},
      archivePrefix={arXiv},
      primaryClass={stat.ML}
}

@misc{ECL,
  author={Trindade,Artur},
  title={{ElectricityLoadDiagrams20112014}},
  year={2015},
  howpublished={UCI Machine Learning Repository},
  note= {{DOI}:https://doi.org/10.24432/C58C86}
}

@article{convtimenet,
  title={ConvTimeNet: A Deep Hierarchical Fully Convolutional Model for Multivariate Time Series Analysis},
  author={Cheng, Mingyue and Yang, Jiqian and Pan, Tingyue and Liu, Qi and Li, Zhi},
  journal={arXiv preprint arXiv:2403.01493},
  year={2024}
}

@misc{onefitsall,
      title={One Fits All:Power General Time Series Analysis by Pretrained LM}, 
      author={Tian Zhou and PeiSong Niu and Xue Wang and Liang Sun and Rong Jin},
      year={2023},
      eprint={2302.11939},
      archivePrefix={arXiv},
      primaryClass={cs.LG}
}

@article{itransformer,
  title={itransformer: Inverted transformers are effective for time series forecasting},
  author={Liu, Yong and Hu, Tengge and Zhang, Haoran and Wu, Haixu and Wang, Shiyu and Ma, Lintao and Long, Mingsheng},
  journal={arXiv preprint arXiv:2310.06625},
  year={2023}
}

@inproceedings{moderntcn,
  title={ModernTCN: A modern pure convolution structure for general time series analysis},
  author={Luo, Donghao and Wang, Xue},
  booktitle={The Twelfth International Conference on Learning Representations},
  year={2024}
}

@article{tide,
  title={Long-term forecasting with tide: Time-series dense encoder},
  author={Das, Abhimanyu and Kong, Weihao and Leach, Andrew and Sen, Rajat and Yu, Rose},
  journal={arXiv preprint arXiv:2304.08424},
  year={2023}
}

@misc{frets,
      title={Frequency-domain MLPs are More Effective Learners in Time Series Forecasting}, 
      author={Kun Yi and Qi Zhang and Wei Fan and Shoujin Wang and Pengyang Wang and Hui He and Defu Lian and Ning An and Longbing Cao and Zhendong Niu},
      year={2023},
      eprint={2311.06184},
      archivePrefix={arXiv},
      primaryClass={cs.LG}
}

@misc{yang2024uncovering,
      title={Uncovering Selective State Space Model's Capabilities in Lifelong Sequential Recommendation}, 
      author={Jiyuan Yang and Yuanzi Li and Jingyu Zhao and Hanbing Wang and Muyang Ma and Jun Ma and Zhaochun Ren and Mengqi Zhang and Xin Xin and Zhumin Chen and Pengjie Ren},
      year={2024},
      eprint={2403.16371},
      archivePrefix={arXiv},
      primaryClass={cs.IR}
}

@misc{film,
      title={FiLM: Frequency improved Legendre Memory Model for Long-term Time Series Forecasting}, 
      author={Tian Zhou and Ziqing Ma and Xue wang and Qingsong Wen and Liang Sun and Tao Yao and Wotao Yin and Rong Jin},
      year={2022},
      eprint={2205.08897},
      archivePrefix={arXiv},
      primaryClass={cs.LG}
}

@inproceedings{revin,
  title={Reversible instance normalization for accurate time-series forecasting against distribution shift},
  author={Kim, Taesung and Kim, Jinhee and Tae, Yunwon and Park, Cheonbok and Choi, Jang-Ho and Choo, Jaegul},
  booktitle={International Conference on Learning Representations},
  year={2021}
}

@misc{liu2024mamba4rec,
      title={Mamba4Rec: Towards Efficient Sequential Recommendation with Selective State Space Models}, 
      author={Chengkai Liu and Jianghao Lin and Jianling Wang and Hanzhou Liu and James Caverlee},
      year={2024},
      eprint={2403.03900},
      archivePrefix={arXiv},
      primaryClass={cs.IR}
}

@article{scinet,
  title={Scinet: Time series modeling and forecasting with sample convolution and interaction},
  author={Liu, Minhao and Zeng, Ailing and Chen, Muxi and Xu, Zhijian and Lai, Qiuxia and Ma, Lingna and Xu, Qiang},
  journal={Advances in Neural Information Processing Systems},
  volume={35},
  pages={5816--5828},
  year={2022}
}

@misc{mamba,
      title={Mamba: Linear-Time Sequence Modeling with Selective State Spaces}, 
      author={Albert Gu and Tri Dao},
      year={2023},
      eprint={2312.00752},
      archivePrefix={arXiv},
      primaryClass={cs.LG}
}

@misc{dlinear,
      title={Are Transformers Effective for Time Series Forecasting?}, 
      author={Ailing Zeng and Muxi Chen and Lei Zhang and Qiang Xu},
      year={2022},
      eprint={2205.13504},
      archivePrefix={arXiv},
      primaryClass={cs.AI}
}

@misc{preformer,
      title={Preformer: Predictive Transformer with Multi-Scale Segment-wise Correlations for Long-Term Time Series Forecasting}, 
      author={Dazhao Du and Bing Su and Zhewei Wei},
      year={2022},
      eprint={2202.11356},
      archivePrefix={arXiv},
      primaryClass={cs.LG}
}

@article{fredf,
  title={FreDF: Learning to Forecast in Frequency Domain},
  author={Wang, Hao and Pan, Licheng and Chen, Zhichao and Yang, Degui and Zhang, Sen and Yang, Yifei and Liu, Xinggao and Li, Haoxuan and Tao, Dacheng},
  journal={arXiv preprint arXiv:2402.02399},
  year={2024}
}

@article{fits,
  title={FITS: Modeling Time Series with $10 k $ Parameters},
  author={Xu, Zhijian and Zeng, Ailing and Xu, Qiang},
  journal={arXiv preprint arXiv:2307.03756},
  year={2023}
}

@inproceedings{timesnet,
  title={Timesnet: Temporal 2d-variation modeling for general time series analysis},
  author={Wu, Haixu and Hu, Tengge and Liu, Yong and Zhou, Hang and Wang, Jianmin and Long, Mingsheng},
  booktitle={The eleventh international conference on learning representations},
  year={2022}
}

@inproceedings{fedformer,
  title={Fedformer: Frequency enhanced decomposed transformer for long-term series forecasting},
  author={Zhou, Tian and Ma, Ziqing and Wen, Qingsong and Wang, Xue and Sun, Liang and Jin, Rong},
  booktitle={International conference on machine learning},
  pages={27268--27286},
  year={2022},
  organization={PMLR}
}

@misc{ftmixer,
      title={FTMixer: Frequency and Time Domain Representations Fusion for Time Series Modeling}, 
      author={Zhengnan Li and Yunxiao Qin and Xilong Cheng and Yuting Tan},
      year={2024},
      eprint={2405.15256},
      archivePrefix={arXiv},
      primaryClass={cs.LG},
      url={https://arxiv.org/abs/2405.15256}, 
}

@inproceedings{crossformer,
  title={Crossformer: Transformer utilizing cross-dimension dependency for multivariate time series forecasting},
  author={Zhang, Yunhao and Yan, Junchi},
  booktitle={The eleventh international conference on learning representations},
  year={2022}
}

@article{wu2021autoformer,
  title={Autoformer: Decomposition transformers with auto-correlation for long-term series forecasting},
  author={Wu, Haixu and Xu, Jiehui and Wang, Jianmin and Long, Mingsheng},
  journal={Advances in neural information processing systems},
  volume={34},
  pages={22419--22430},
  year={2021}
}

@Inbook{simplex,
author="Eaves, B. Curtis",
title="Standard Simplex S and Matrix Operations",
bookTitle="A Course in Triangulations for Solving Equations with Deformations",
year="1984",
publisher="Springer Berlin Heidelberg",
address="Berlin, Heidelberg",
pages="39--44",
isbn="978-3-642-46516-1",
doi="10.1007/978-3-642-46516-1_4",
url="https://doi.org/10.1007/978-3-642-46516-1_4"
}

@misc{yi2024filternetharnessingfrequencyfilters,
      title={FilterNet: Harnessing Frequency Filters for Time Series Forecasting}, 
      author={Kun Yi and Jingru Fei and Qi Zhang and Hui He and Shufeng Hao and Defu Lian and Wei Fan},
      year={2024},
      eprint={2411.01623},
      archivePrefix={arXiv},
      primaryClass={cs.LG},
      url={https://arxiv.org/abs/2411.01623}, 
}

@article{rademacher,
author = {Bartlett, Peter L. and Mendelson, Shahar},
title = {Rademacher and gaussian complexities: risk bounds and structural results},
year = {2003},
issue_date = {3/1/2003},
publisher = {JMLR.org},
volume = {3},
number = {null},
issn = {1532-4435},
journal = {J. Mach. Learn. Res.},
month = mar,
pages = {463–482},
numpages = {20}
}

@misc{chen2023tsmixerallmlparchitecturetime,
      title={TSMixer: An All-MLP Architecture for Time Series Forecasting}, 
      author={Si-An Chen and Chun-Liang Li and Nate Yoder and Sercan O. Arik and Tomas Pfister},
      year={2023},
      eprint={2303.06053},
      archivePrefix={arXiv},
      primaryClass={cs.LG},
      url={https://arxiv.org/abs/2303.06053}, 
}

@misc{fredo,
      title={FreDo: Frequency Domain-based Long-Term Time Series Forecasting}, 
      author={Fan-Keng Sun and Duane S. Boning},
      year={2022},
      eprint={2205.12301},
      archivePrefix={arXiv},
      primaryClass={cs.LG},
      url={https://arxiv.org/abs/2205.12301}, 
}

@String{Computing = "Computing" }

@String{Computer = "{IEEE} Computer" }

@String{Springer = "Springer-Verlag" }

@ArtifactSoftware{R,
    title = {R: A Language and Environment for Statistical Computing},
    author = {{R Core Team}},
    organization = {R Foundation for Statistical Computing},
    address = {Vienna, Austria},
    year = {2019},
    url = {https://www.R-project.org/},
}
\bibliographystyle{ieeetr}
\end{document}